\documentclass[11pt, letterpaper]{cmu}

\usepackage[all]{hypcap}
\usepackage[comma,numbers,sort,compress]{natbib}
\usepackage{hyperref}[citecolor=magenta]
\usepackage{amsmath, amssymb}

\hypersetup{
    colorlinks = true,
    citecolor = {magenta},
}

    \PassOptionsToPackage{numbers, compress}{natbib}

\usepackage[utf8]{inputenc} 
\usepackage[T1]{fontenc}    
\usepackage{url}            
\usepackage{booktabs}       
\usepackage{amsfonts}       
\usepackage{nicefrac}       
\usepackage{microtype}     %
\usepackage{xcolor}         
\microtypecontext{spacing=nonfrench}
\usepackage{amsmath,amsthm}
\usepackage{graphicx}

\usepackage{mathtools}
\usepackage{mathrsfs}
\usepackage{dsfont}
\usepackage{float}

\usepackage{amssymb}
\usepackage{pifont}

\usepackage{xspace}
\usepackage[capitalize,noabbrev]{cleveref}
\usepackage{lipsum}
\usepackage{listings}

\usepackage{bbm}
\usepackage{tabularx}
\usepackage{lmodern}
\usepackage[T1]{fontenc}
\usepackage{lmodern}
\usepackage{amsmath, amssymb}
\usepackage{multirow}
\usepackage[export]{adjustbox}
\usepackage{colortbl}

\usepackage{algpseudocode}
\usepackage{setspace}
\usepackage{booktabs}

\usepackage{color}
\definecolor{deepblue}{rgb}{0,0,0.5}
\definecolor{deepred}{rgb}{0.6,0,0}
\definecolor{deepgreen}{rgb}{0,0.5,0}
\definecolor{boost_correct_to_correct}{HTML}{66C2A5}
\definecolor{default_correct_to_correct}{HTML}{fc8d62}
\definecolor{dup_correct_to_correct}{HTML}{8da0cb}
\definecolor{new_correct_to_correct}{HTML}{e78ac3}

\usepackage{tikz}

\definecolor{lightbluez}{RGB}{187, 214, 238}
\definecolor{lightorangez}{RGB}{247, 202, 172}

\newcommand{\hlblue}[1]{%
  \tikz[baseline=(X.base)]\node[rectangle, fill=lightbluez!50, inner sep=2pt, outer sep=0pt](X){$#1$};%
}
\newcommand{\hlorange}[1]{%
  \tikz[baseline=(X.base)]\node[rectangle, fill=lightorangez!50, inner sep=2pt, outer sep=0pt](X){$#1$};%
}

\usepackage[most,skins,theorems]{tcolorbox}

\theoremstyle{plain}

\theoremstyle{definition}

\theoremstyle{remark}

\usepackage{amsmath,amsfonts,bm,scalefnt}

\definecolor{lightblue}{rgb}{0.22,0.50,0.70}
\def\eqref#1{Eq.~\ref{#1}}

\def\1{\bm{1}}

\DeclareMathAlphabet{\mathsfit}{\encodingdefault}{\sfdefault}{m}{sl}
\SetMathAlphabet{\mathsfit}{bold}{\encodingdefault}{\sfdefault}{bx}{n}

\usepackage{wrapfig}
\usepackage{subcaption}

\definecolor{rliableolive}{HTML}{BBCC33}
\definecolor{rliableblue}{HTML}{77AADD}
\definecolor{rliablered}{HTML}{EE8866}
\definecolor{figurebackground}{HTML}{F9F8F3}

\tcbset{
  aibox/.style={
    width=\linewidth,
    top=6pt,
    bottom=0pt,
    left=1.5pt,
    right=1.5pt,
    colback=rliableolive!13!white,
    colframe=black,
    colbacktitle=black,
    enhanced,
    center,
    attach boxed title to top left={yshift=-0.1in,xshift=0.15in},
    boxed title style={boxrule=0pt,colframe=white,},
  }
}
\newtcolorbox{AIbox}[2][]{aibox,title=#2,#1}
\usepackage{enumitem}

\newcommand\pythonstyle{\lstset{
basicstyle=\ttfamily\footnotesize,
language=Python,
morekeywords={self, clip, exp, mse_loss, uniform_sample, concatenate, logsumexp},              
keywordstyle=\color{deepblue},
emph={MyClass,__init__},          
emphstyle=\color{deepred},   
stringstyle=\color{deepgreen},
frame=single,                       
showstringspaces=false
}}

\lstnewenvironment{python}[1][]
{
\pythonstyle
\lstset{#1}
}
{}

\newcommand\pythoninline[1]{{\pythonstyle\lstinline!#1!}}

\newcommand{\x}{\mathbf{x}}
\newcommand{\z}{\mathbf{z}}
\newcommand{\y}{\mathbf{y}}

\newcommand{\methodname}{\texttt{\textbf{RC}}}

\definecolor{blanchedalmond}{rgb}{1.0, 0.92, 0.8}
\definecolor{carmine}{rgb}{0.59, 0.0, 0.09}
\definecolor{lightblue}{rgb}{0.22,0.45,0.70}

\renewcommand{\mathbf}{\boldsymbol}

\makeatletter
\def\Ddots{\mathinner{\mkern1mu\raise\p@
\vbox{\kern7\p@\hbox{.}}\mkern2mu
\raise4\p@\hbox{.}\mkern2mu\raise7\p@\hbox{.}\mkern1mu}}
\makeatother

\numberwithin{equation}{section}

\definecolor{amaranth}{rgb}{0.9, 0.17, 0.31}
\definecolor{antiquebrass}{rgb}{0.8, 0.58, 0.46}
\definecolor{antiquefuchsia}{rgb}{0.57, 0.36, 0.51}
\definecolor{chromeyellow}{rgb}{0.31, 0.47, 0.26}

\usepackage[dvipsnames]{xcolor}
\definecolor{maj5}{HTML}{2b8cbe}
\definecolor{maj5Imp}{HTML}{084081}

\definecolor{seq5wo}{HTML}{d95f0e}
\definecolor{seq5woImp}{HTML}{662506}

\definecolor{seq5w}{HTML}{6a51a3}
\definecolor{seq5wImp}{HTML}{3f007d}

\definecolor{selfwo}{HTML}{d95f0e}
\definecolor{selfwoImp}{HTML}{662506}

\definecolor{selfw}{HTML}{6a51a3}
\definecolor{selfwImp}{HTML}{3f007d}

\definecolor{glorewo}{HTML}{d95f0e}
\definecolor{glorewoImp}{HTML}{662506}

\definecolor{glorew}{HTML}{6a51a3}
\definecolor{glorewImp}{HTML}{3f007d}

\definecolor{vstar}{HTML}{d95f0e}
\definecolor{vstarImp}{HTML}{662506}

\makeatletter
\def\mathcolor#1#{\@mathcolor{#1}}
\def\@mathcolor#1#2#3{%
  \protect\leavevmode
  \begingroup
    \color#1{#2}#3%
  \endgroup
}
\makeatother

\usepackage[textsize=tiny]{todonotes}
\usepackage{algorithm}
\usepackage{amssymb}

\usepackage[skins,theorems]{tcolorbox}
\Crefformat{equation}{#2Eq.\;(#1)#3}

\Crefformat{figure}{#2Figure #1#3}
\Crefformat{assumption}{#2Assumption #1#3}
\Crefname{assumption}{Assumption}{Assumptions}

\usepackage{crossreftools}
\usepackage[suppress]{color-edits}

\usepackage{dsfont}
\usepackage{nicefrac}

\newtcolorbox{analysisbox}[1][]{
    enhanced jigsaw,
    colback=white,
    colframe=blue!75!black,
    fonttitle=\bfseries,
    boxsep=5pt,
    left=5pt,
    right=5pt,
    top=5pt,
    bottom=5pt,
    title=#1,
}

\tcbset{
  aibox/.style={
    width=\linewidth,
    top=8pt,
    bottom=4pt,
    colback=rliableolive!13!white,
    colframe=black,
    colbacktitle=black,
    enhanced,
    center,
    attach boxed title to top left={yshift=-0.1in,xshift=0.15in},
    boxed title style={boxrule=0pt,colframe=white,},
  }
}
\definecolor{lightblue}{rgb}{0.22,0.45,0.70}

\usepackage{pifont}
\usepackage{soul}
\usepackage{fancyvrb}

\definecolor{highlightmistake}{RGB}{255, 179, 179}
\definecolor{highlightcorrect}{RGB}{179, 255, 179}

\title{Reasoning Cache: Continual Improvement Over Long Horizons via Short-Horizon RL}

\author[1]{Ian Wu}
\author[1]{Yuxiao Qu}
\author[1]{Amrith Setlur}
\author[1]{Aviral Kumar}

\affil[1]{Carnegie Mellon University}

\correspondingauthor{ianwu@andrew.cmu.edu}
\coderepository{github.com/IanYHWu/rc}

\begin{document}

\maketitle

\begin{figure}[htbp]
\vspace{-1.2cm}
    \centering
    \begin{subfigure}[c]{0.67\textwidth}
     \captionsetup{font=small,skip=2pt}
        \centering
        \includegraphics[width=0.99\textwidth]{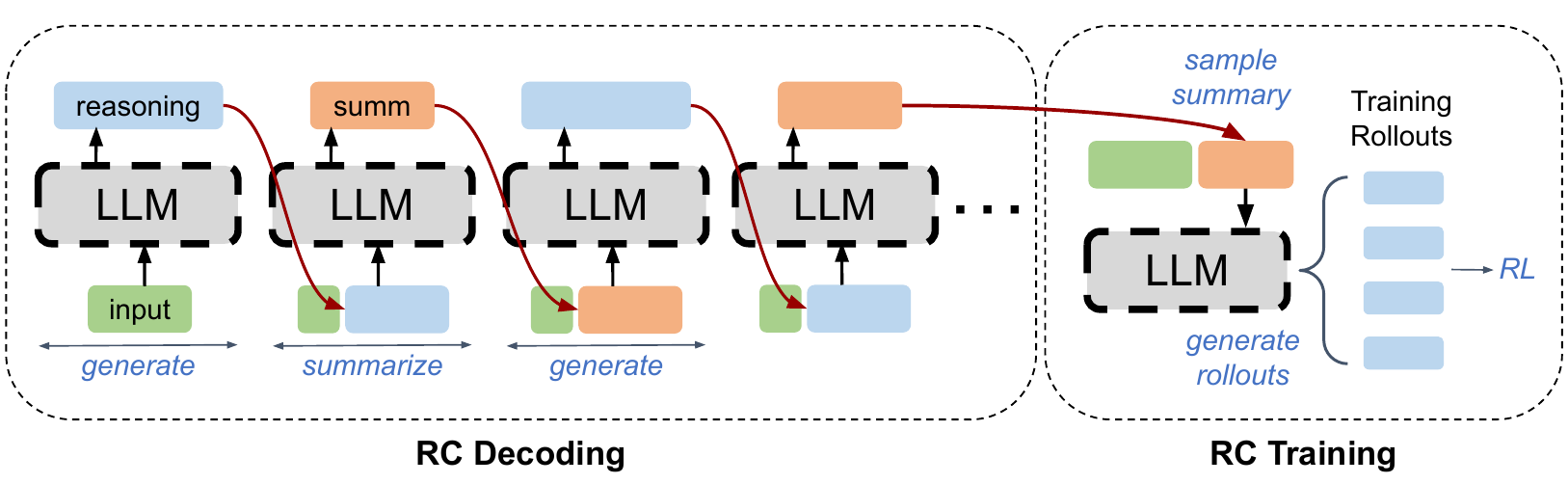}
        \caption*{}
    \end{subfigure}\hfill
    \begin{subfigure}[c]{0.31\textwidth}
     \captionsetup{font=small,skip=2pt}
        \centering
        \vspace{0.2cm} \includegraphics[width=0.99\textwidth]{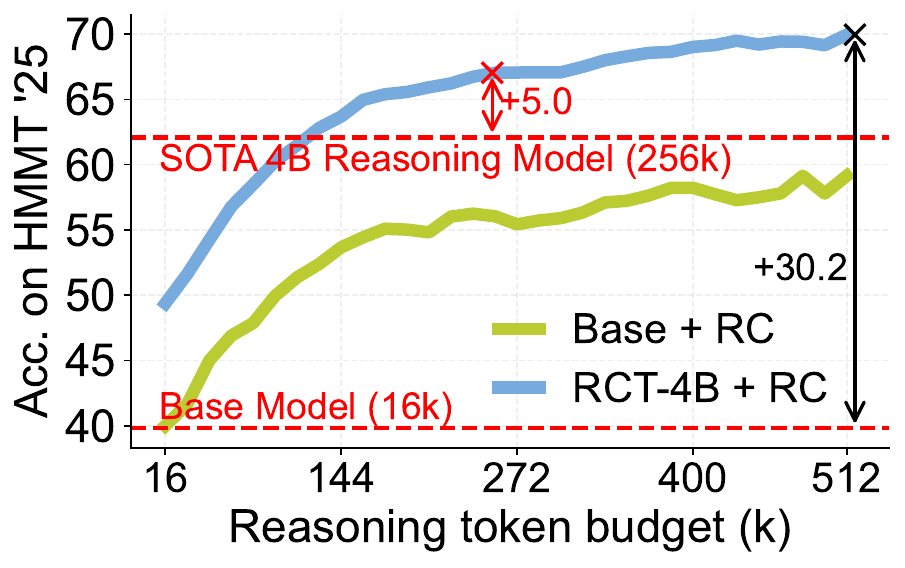}
        \caption*{}
    \end{subfigure}
    \vspace{-0.5cm}
    \caption{\footnotesize{\textbf{\textit{Left:} Illustration of the \methodname{} algorithm.} \methodname{} decoding replaces standard autoregressive decoding at both train and test time. During \methodname{} decoding, the LLM generates a reasoning trace, summarizes it, discards the original trace, and conditions subsequent reasoning on this summary. This design decouples the effective reasoning horizon from the length of any single reasoning trace, thus maintaining tractable rollout lengths for outcome-reward RL while also enabling extrapolation at test time. \textbf{\textit{Right:} Performance on HMMT 2025 (November) vs. reasoning token budget.} Our \methodname{}-trained model \texttt{RCT-4B} (blue, trained from Qwen3-4B-Instruct-2507 at 16k train budget) extrapolates to outperform both the base model with RC decoding (green) and the specialized Qwen3-4B-Thinking-2507 reasoning model (evaluated at 256k test tokens).
    }}
    \label{fig:main_fig}
    \vspace{-0.1cm}
\end{figure}

{\absfont \textbf{Abstract:} Large Language Models (LLMs) that can continually improve beyond their training budgets are able to solve increasingly difficult problems by adapting at test time, a property we refer to as \emph{extrapolation}. However, standard reinforcement learning (RL) operates over fixed problem distributions and training budgets, which limits extrapolation amidst distribution shift at test time. To address this, we introduce \methodname{}, an iterative decoding algorithm that replaces autoregressive decoding during both training and inference. \methodname{} exploits an \emph{asymmetry} between the response generation and summarization capabilities of LLMs to construct reasoning chains that consistently improve across iterations. Models trained to use \methodname{} can extrapolate and improve over reasoning horizons more than an order of magnitude longer than those seen during training. Empirically, training a 4B model with \methodname{} using a 16k-token training budget improves performance on HMMT 2025 from 40\% to nearly 70\% when evaluated with 0.5m test-time tokens, outperforming both comparably sized models and many larger reasoning LLMs. We also show that models trained with \methodname{} can more effectively leverage existing scaffolds to further scale test-time performance, due to the improved summary-conditioned generation abilities learned through training.}

\vspace{-0.25cm}
\section{Introduction}
\label{sec:introduction}
\vspace{-0.2cm}

Large language models (LLMs) exhibit the ability to solve complex problems by generating long reasoning traces at test time. As LLMs become more capable, we naturally expect them to be able to solve harder tasks by reasoning for longer, even without external supervision. This expectation mirrors human cognition: humans continually improve their reasoning by revisiting earlier conclusions and reallocating effort to discover new information in new ways over the course of problem solving. We would like our models to behave similarly, such that additional reasoning at test time translates to improved performance over horizons much longer than any single reasoning step. This underlies the notion of \emph{in-context exploration}~\citep{setlur2025e3learningexploreenables}, which learns an implicit algorithm or ``procedure'' for allocating test-time compute such that spending more computation systematically improves outcomes. If learned robustly, in-context exploration should enable models to continually improve over long horizons at test time, perhaps over hours or even days, and across millions of tokens, all without modifying the model weights.

However, current LLM training paradigms limit the forms of in-context exploration that can be learned. Supervised fine-tuning (SFT) teaches models to imitate the content of reasoning traces, rather than learning the algorithm that generates them~\citep{chu2025sftmemorizesrlgeneralizes,setlur2025scaling}. This inhibits the learning of systematic reasoning procedures and thus the ability to perform effective in-context exploration. While reinforcement learning (RL) incentivizes models to learn reasoning procedures rather than mere imitation~\citep{sun2025rlgrokkingrecipedoes, zhang2025interplaypretrainingmidtrainingrl}, RL is typically applied over a fixed prompt distribution and a bounded training rollout length. As a result, models are optimized to exploit this finite budget, rather than to \emph{extrapolate} beyond it. When such models encounter complex problems that require more reasoning to solve, two failure modes emerge. First, they may prematurely terminate within their training budgets and fail to make progress by reasoning for longer. Second, when models continue beyond this budget, \emph{distribution shift} may occur as generation proceeds from conditional distributions that differ substantially from those seen during training; reasoning traces in this regime are often repetitive and verbose~\citep{deepscaler2025, setlur2025e3learningexploreenables}. This raises a natural question: \emph{\textbf{how can we \emph{train} models to extrapolate their reasoning far beyond their training configurations?}}

To answer this question, we introduce \textbf{\emph{Reasoning Cache}} (\methodname), an iterative decoding algorithm that replaces standard autoregressive decoding during both training and inference. In \methodname{}, the model generates a reasoning trace, summarizes it (the ``cache'') and discards the original trace over multiple turns, with subsequent generation conditioning only on the previous summary rather than the full reasoning trace. For training, we introduce an RL approach that teaches the model to exploit \methodname{} by improving its summary-conditioned reasoning capabilities. Our approach combines typical on-policy RL with off-policy learning via a \emph{replay buffer} that enables reuse of cached summaries, allowing the model to train over long effective horizons without generating prohibitively long training rollouts. The design of the decoding process used by \methodname{} is motivated by two key observations. First, iterative decoding provides explicit control over test-time compute: we can scale the number of reasoning iterations while keeping each step within the training distribution, mitigating distribution shift as effective reasoning length grows. Second, making such iteration effective requires consistent progress across steps. \methodname{} achieves this by exploiting a \emph{summarization–generation asymmetry}: models are often better at summarizing prior reasoning and reasoning from summaries than at producing correct solutions from scratch. Our RL objective explicitly amplifies this asymmetry, enabling larger improvements across successive iterations.

Empirically, models trained with \methodname{} exhibit strong and consistent extrapolation ability. While \methodname{} decoding alone is already effective when the base model can reliably follow instructions and reason from summaries, \methodname{}-trained models extrapolate substantially further. On the mathematical reasoning benchmarks HMMT 2025 (November) and IMO-AnswerBench, our \methodname{}-trained \texttt{RCT-4B} model substantially outperforms both Qwen3-4B-Instruct-2507 (the base model) and Qwen3-4B-Thinking-2507 (a much stronger reasoning model) through extrapolation. For example, accuracy on HMMT 2025 improves from 40\% to 70\% as the test budget is scaled from 16k to 512k tokens (Figure~\ref{fig:main_fig}, right), while on IMO-AnswerBench~\citep{luong2025robustmathematicalreasoning}, performance rises from 41\% at 16k tokens to 58\% at 256k tokens, surpassing larger models such as Qwen3-235B, Qwen3-30B-A3B-Instruct-2507 and Nemotron-3-Nano-30B-A3B using standard autoregressive decoding, despite training only at a 16k budget.
Notably, despite having been trained solely on mathematical reasoning data, \texttt{RCT-4B} also achieves substantially higher performance than the base model on the FrontierScience~\citep{openai2025frontierscience} scientific reasoning benchmark when extrapolated to 256k tokens, suggesting that \methodname{} induces transferable algorithmic behavior rather than domain-specific knowledge.
Finally, we show that \texttt{RCT-4B} is consistently better at exploiting a variety of test-time scaffolds than both the base model and a base model post-trained with standard RL, with and without \methodname{} decoding. This indicates that optimizing for summary-conditioned generation yields models that are not only better at leveraging \methodname{} decoding, but are also generally more effective at using external context to guide reasoning.

\vspace{-0.25cm}
\section{Related Work}
\label{sec:related_work}
\vspace{-0.2cm}

\textbf{Long-horizon reasoning by scaling test-time compute.} Increasing the length of reasoning traces is a fundamental approach to scaling test-time compute~\citep{wei2019regularization, Guo_2025}, as longer traces allow for more extensive in-context exploration that can substantially improve performance~\citep{setlur2025e3learningexploreenables, gandhi2025cognitivebehaviorsenableselfimproving, zhang2025interplaypretrainingmidtrainingrl}. Although prior work finds that on-policy RL is effective at teaching LLMs to generate longer autoregressive responses~\citep{Guo_2025}, scaling in this way suffers from a fundamental limitation: models cannot reliably extrapolate much beyond the reasoning lengths seen during RL training~\citep{deepscaler2025, setlur2025e3learningexploreenables}. Our work therefore aims to develop methods that enable extrapolation, rather than merely scaling reasoning lengths by increasing training budgets.
 
\textbf{Test-time extrapolation of reasoning.} Prior work attempts to enable extrapolation mainly through one of two approaches. The first modifies training via carefully designed datasets and curricula to encourage in-context exploration~\citep{setlur2025e3learningexploreenables, Polaris2025, deepscaler2025}. Although this has been shown to enable extrapolation to around $3\text{--}4\times$ the training budget, performance typically saturates beyond this. The second approach modifies the RL reward structure to implicitly encourage extrapolation (e.g., via dense rewards that credit intermediate reasoning segments~\citep{qu2025optimizingtesttimecomputemeta}) but retains pure autoregressive decoding at inference time. In both cases, in-context exploration behaviors are implicitly learned through free-form autoregressive generation and thus remain coupled to length budgets in the training setup. When test-time conditional distributions fall outside the training support, as is seen when test lengths greatly exceed train lengths, these approaches do not improve further and instead result in verbose and repetitive behavior~\citep{setlur2025e3learningexploreenables, deepscaler2025}.

\textbf{Iterative decoding for scaling test-time compute.} Prior work explores prompting LLMs to iteratively transform their own outputs to scale test-time compute, ranging from simple self-correction~\citep{huang2024largelanguagemodelsselfcorrect, kim2023languagemodelssolvecomputer} and self-refinement~\citep{madaan2023self, shinn2023reflexion} to more complex scaffolds that combine iterative and parallel compute~\citep{ shao2025deepseekmathv2selfverifiablemathematicalreasoning, venkatraman2025recursiveselfaggregationunlocksdeep}. Others consider training models to apply transformations more effectively rather than relying on prompting alone~\citep{qu2024recursiveintrospectionteachinglanguage, kumar2024traininglanguagemodelsselfcorrect}. Such transformations may occur over multiple time steps: \citet{venkatraman2025recursiveselfaggregationunlocksdeep}, for example, train models to iteratively aggregate parallel reasoning traces, 
while \citet{madaan2025rethinkingthinkingtokensllms} consider distilling parallel reasoning traces into summaries that are conditioned on for final answer generation. In contrast to these works, \methodname{} scales test-time compute sequentially through extrapolation rather than in parallel. This is reflected in our training procedure, which directly optimizes for multi-step reasoning that can extend to very long horizons through the use of a summary replay buffer. Furthermore, we find that \methodname{} training enables models to better exploit other scaffolds by teaching them to more effectively leverage external guidance, thus making our approach complementary to many existing works.

\textbf{Memory in multi-turn interaction.} \methodname{} summaries can be viewed as compressed memory states that are updated as the policy acts over iterations. Prior work primarily consider using similar memory states to store external context (e.g. retrieved web pages, user responses etc.) that is dynamically recalled in later steps, often as part of multi-turn question-answering or conversation systems~\citep{li2023compressingcontextenhanceinference, zhou2025mem1learningsynergizememory}. Our method instead uses memory states to store abstractions of self-generated reasoning traces for solving reasoning problems, an approach that has been explored using prompting-based approaches~\citep{suzgun2025dynamiccheatsheettesttimelearning, ho2025arcmemoabstractreasoningcomposition, wei2025evomemorybenchmarkingllmagent}. Unlike these works, we focus on training the model to better utilize these memory states for downstream reasoning, which we show yields significant improvements over  prompting-only methods.
\vspace{-0.25cm}
\section{Preliminaries and Notation}
\label{sec:analysis}
\vspace{-0.2cm}

Consider a policy $\pi_\theta(\cdot|\x)$ that generates tokens autoregressively conditioned on $\x$. At test time, the policy is given a token budget and allocates this to reason. Our interest is test-time performance as a function of the test token budget, particularly with large test-time budgets.

\textbf{Standard RL training for LLM reasoning.} Let $\mathcal{D}_{\mathrm{train}}$ denote a training distribution of prompt-answer pairs $(\x, \y)$. On-policy reinforcement learning (RL) optimizes the expected reward of the policy:
\vspace{-0.2cm}
\begin{align}
\label{eq:standard_rl_training}
    \max_{\pi_\theta}~~ 
    \mathbb{E}_{\x, \y \sim \mathcal{D}_{\text{train}}} 
    \left[ 
    \mathbb{E}_{\z \sim \pi_{\theta}(\cdot \mid \x)} 
    [r(\y, \z)] 
    \right],\quad
    ~~\text{s.t.}~~ |\z| \leq H_{\mathrm{train}}.~~~~~ \text{(Training objective)}
\end{align}
Here, $\z$ denotes an on-policy rollout autoregressively sampled from $\pi_\theta$. The rollout encodes a reasoning trace and is generated within a fixed training budget $H_{\mathrm{train}}$. The reward function $r(\y, \z)$ evaluates the correctness of the rollout, typically by extracting the final answer from $\z$ and comparing it against the ground-truth label $\y$. To solve this optimization problem, we can use outcome-reward policy-gradient methods: one common choice is GRPO~\citep{shao2024deepseekmathpushinglimitsmathematical} (see Appendix~\ref{app:grpo_overview}), which we use throughout this work.

\textbf{Test-time extrapolation of LLM reasoning.} Equation~\ref{eq:standard_rl_training} optimizes performance only over the empirical distribution of training prompts $\mathcal{D}_{\text{train}}$, and only under a fixed  $H_{\text{train}}$. At test time, we may want to ensure that our trained model attains high accuracy on a different prompt distribution and under a larger budget:
\begin{align}
\label{eq:standard_rl_testing}
    \mathrm{TestPerf}(\pi_\theta) \overset{\mathrm{def}}{:=} 
    \mathbb{E}_{\x, \y \sim \mathcal{D}_{\text{test}}} 
    \left[ 
    \mathbb{E}_{\z \sim \pi_{\theta}(\cdot \mid \x)} 
    [r(\y, \z)] 
    \right],\quad
    \text{s.t.}~~ |\z| \leq H_{\mathrm{test}} \quad \text{(Test-time objective)}
\end{align}
where $H_\text{test}$ is the test budget; in general, the training and test distributions differ (i.e. $p_\text{train}(\x) \neq p_\text{test}(\x)$ and $H_\text{test} \gg H_\text{train}$). When a model trained to optimize Equation~\ref{eq:standard_rl_training} can leverage a larger test budget to achieve $\mathrm{TestPerf}(\pi_\theta)|_{H_\text{test}} > \mathrm{TestPerf}(\pi_\theta)|_{H_\text{train}}$, we say that it can \emph{extrapolate}.

\vspace{-0.25cm}
\section{Problem Statement}
\label{sec:problem_statement}
\vspace{-0.2cm}

\textbf{\emph{Does optimizing performance at $H_\text{train}$ (Equation~\ref{eq:standard_rl_training}) also optimize extrapolation at $H_\text{test}$ (Equation~\ref{eq:standard_rl_testing})?}} Unfortunately, the answer is \textbf{no}. During training, the model receives positive reward only for rollouts that terminate within $H_\text{train}$ tokens. This implicitly penalizes longer rollouts and encourages ``premature'' termination near $H_\text{train}$ at test time. Moreover, when the model does continue beyond $H_{\text{train}}$ at test time, it must effectively operate on the sorts of conditional distributions it was never trained on. While this form of \emph{distribution shift} is not problematic if the model has learned a true ``operator''~\citep{qu2025optimizingtesttimecomputemeta} that enables the chaining of behaviors to solve problems, it is unclear whether RL can learn such operators from a finite, fixed prompt set. We instead circumvent this challenge by altering the decoding algorithm such that the model never encounters significant distribution shift even when reasoning at long horizons.

\textbf{Why do we need extrapolation?} Can we simply increase $H_{\text{train}}$ to match $H_{\text{test}}$ during RL training? Doing so would obviate extrapolation altogether. However, there are two main problems with this approach. First, any new test distribution we encounter may contain harder problems requiring $H_{\text{test}} \gg H_{\mathrm{train}}$ to solve, and so it would be better to train models that can adapt to larger test budgets on-the-fly (i.e., training models to ``continually adapt''). Second, memory and compute costs, as well as the effectiveness of RL training, scale aggressively with response length, making long-horizon on-policy RL prohibitively expensive and challenging. These challenges indicate that we cannot just scale $H_\text{train}$, and should instead train models to extrapolate as more compute is provided. 

\vspace{-0.25cm}
\section{Enabling Extrapolation with Reasoning Cache}
\label{sec:reasoning_cache}
\vspace{-0.2cm}

Our goal is to develop a method that trains models under a fixed token budget and a finite prompt set, while still allowing them to extrapolate beyond the training horizon. To achieve this, we replace autoregressive decoding with an iterative decoding algorithm \texttt{Alg}$(\pi_\theta; \x)$ during training and inference. This algorithm, \texttt{Alg}, leverages the structure of reasoning along with asymmetries present in LLMs to support long-horizon reasoning at test time while remaining amenable to training under a much smaller $H_{\text{train}}$. We begin by concretizing the key desiderata that \texttt{Alg} should satisfy.

\begin{figure*}[t]
    \centering
    \begin{subfigure}[b]{0.99\textwidth}
    \captionsetup{font=small,skip=2pt}
        \centering
        \includegraphics[width=0.99\textwidth]{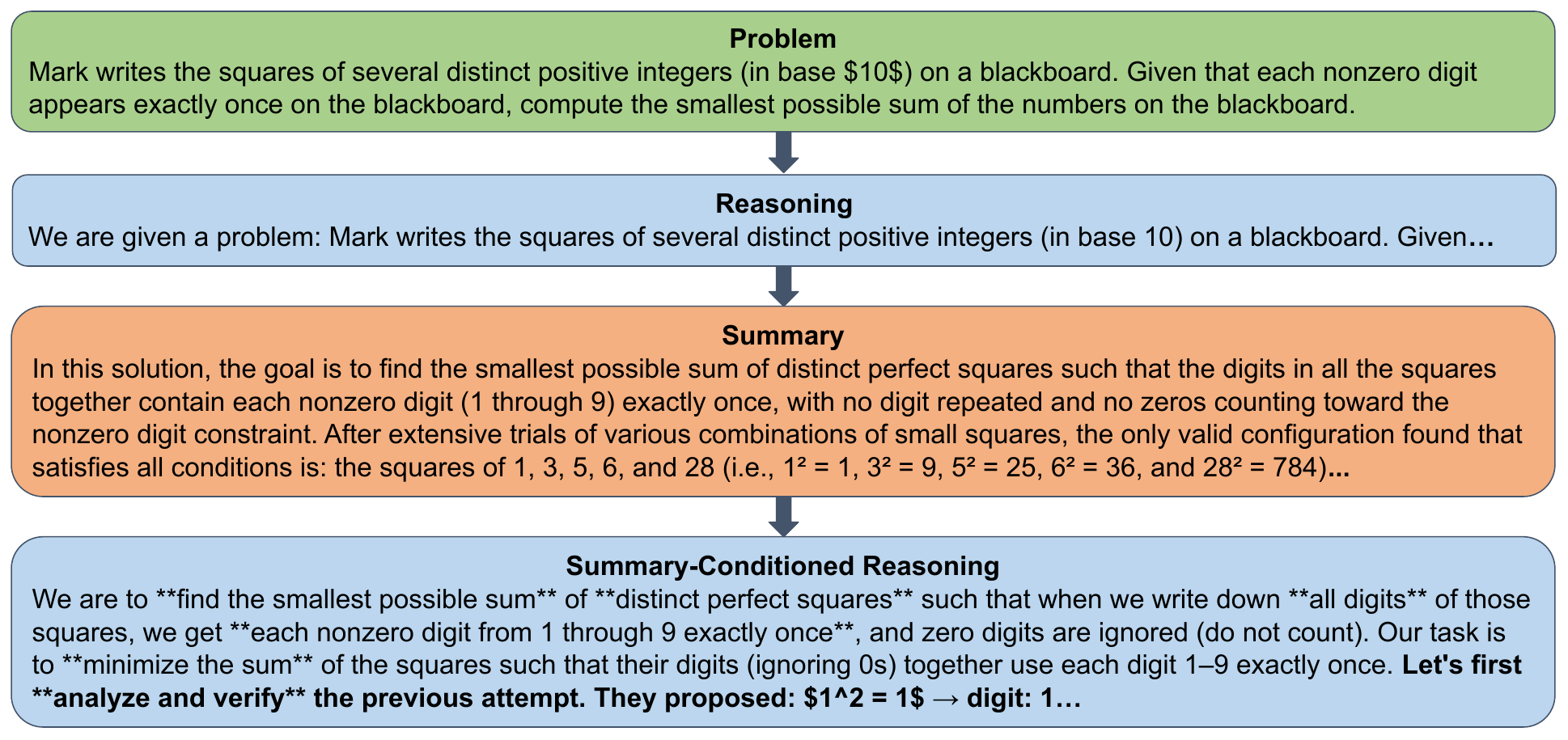}
        \caption*{}
    \end{subfigure}\hfill
    \vspace{-0.6cm}
    \caption{\footnotesize{\textbf{Illustrative example of \methodname{}'s output.} The model generates reasoning given an input problem, before conditioning on the reasoning to generate a summary between 1-2 paragraphs long. The model then conditions on the summary to generate new reasoning. As we show later, \methodname{} training improves the ability of the model to perform \emph{summary-conditioned reasoning}, which enables the model to continually improve over long horizons. See Appendix~\ref{app:add_example} for a full example of an \methodname{} output.}}
    \label{fig:example}
    \vspace{-0.1cm}
\end{figure*}

\textbf{Choosing an effective decoding algorithm.}
An effective choice of \texttt{Alg} must satisfy two desiderata.
\textcolor{lightblue}{\emph{\textbf{First}}}, it should define an iterative procedure in which the number of iterations monotonically controls test-time compute, while each iteration operates on conditional distributions that remain close to those encountered during training.
An \texttt{Alg} satisfying this desideratum avoids the main limitations of autoregressive decoding in standard RL:
\textbf{(1)} by enabling longer responses through increasing the iteration limit, it mitigates premature termination biases induced by fixed-length RL training; and
\textbf{(2)} by restricting autoregressive generation within each iteration to at most $H_{\text{train}}$ tokens, it reduces train–test distribution shift even when the effective reasoning horizon is much larger.
\textcolor{lightblue}{\emph{\textbf{Second}}}, the algorithm should retain expressivity comparable to autoregressive decoding, allowing each iteration to refine or extend the current rollout and explore new directions.
An \texttt{Alg} that satisfies this property can make consistent progress across many iterations, enabling continual extrapolation over long horizons.

\vspace{-0.2cm}
\subsection{\methodname{}: A Multi-Turn Decoding Algorithm}\label{subsec:rc_method}
\vspace{-0.1cm}
We now introduce a decoding algorithm that satisfies these desiderata. Our algorithm, which we call \textbf{\emph{Reasoning Cache}} (\methodname), is an iterative decoding approach that alternates between response generation and summarization. Being an iterative decoding algorithm, \methodname{} naturally fulfills our first desideratum: we can increase test-time compute by increasing the number of summarization-generation turns, while also avoiding significant shifts in the conditional distributions encountered at each turn by only ever autoregressively generating at bounded lengths $H_\text{train} \ll H_\text{test}$. To satisfy our second desideratum, \methodname{} relies on two properties of LLMs. First, reasoning traces contain \emph{redundant} tokens: many tokens encode steps that are useful for local progress but need not be retained verbatim to guide future actions. This allows us to discard a significant portion of tokens (e.g. via summarization) so long as key information is retained. Second, as we consistently find in our experiments, base LLMs often exhibit \emph{summarization-generation asymmetry}, in that producing a correct response conditioned on a summary of a previous attempt is easier than generating a correct response from scratch; this asymmetry arises from the instruction-following abilities of LLMs, which allows them to use summaries of prior generations to guide subsequent reasoning. \methodname{} exploits this by periodically compressing responses into a cache and conditioning subsequent generation on it, allowing the model to refine, extend, or restart reasoning across iterations as needed. See Figure~\ref{fig:example} for an  example of \methodname{}'s outputs.

\begin{figure}[t]
    \centering
    \vspace{-0.2cm}
    \begin{subfigure}[b]{0.99\textwidth}
    \captionsetup{font=small,skip=2pt}
        \centering
        \includegraphics[width=0.99\textwidth]{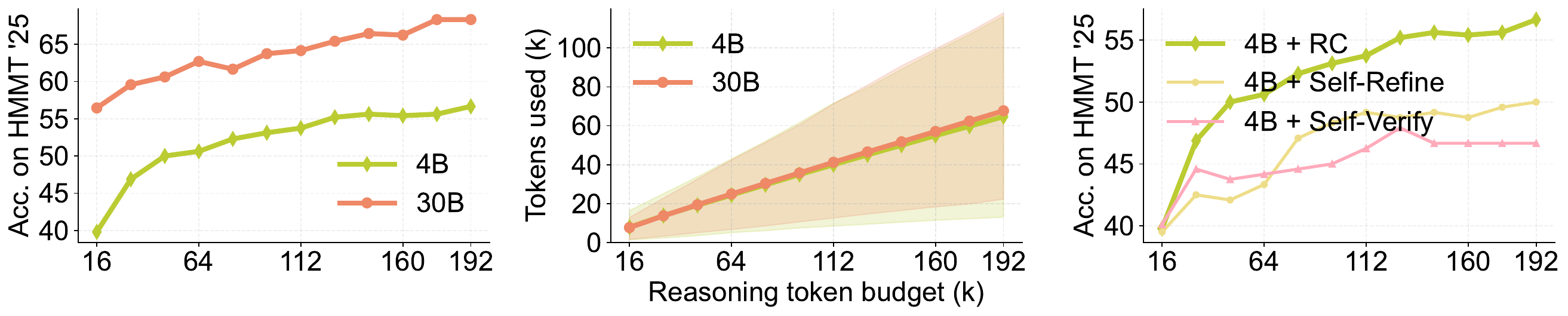}
        \caption*{}
    \end{subfigure}\hfill
    \vspace{-0.6cm}
    \caption{\footnotesize{\textbf{\textit{Left:} Accuracy vs. test-time token budget.} \methodname{} decoding improves performance as token budget $H_\text{test}$ is increased far beyond $H_\text{train} = 16\text{k}$. \textbf{\textit{Middle:} Total tokens used vs. test-time token budget.} Total reasoning tokens used by \methodname{} increases linearly as we increase the reasoning token budget. Shaded regions indicate the 5th-95th percentile; lines indicate the mean. \textbf{\textit{Right:} Accuracy vs. token budget for iterative decoding methods.} \methodname{} is a more effective method for enabling extrapolation than self-verification and self-refinement, highlighting the benefits of exploiting the summarization-generation gap.}}
    \label{fig:rc_analysis_1}
    \vspace{-0.2cm}
\end{figure}

Let $\x$ denote the prompt and let $t \in \mathbb{N}$ index the decoding turn. \methodname{} maintains: \textbf{(1)} a reasoning trace $\z_R^{(\mathrm{t})}$ and \textbf{(2)} a summary $\z_S^{(\mathrm{t})}$, with $\z_S^{(\mathrm{0})}$ initialized to the empty string. At each turn, $\z_R^{(\mathrm{t})}$ is generated under a fixed token budget $H_R$, while $\z_S^{(\mathrm{t})}$ is generated under $H_S \ll H_R$. Decoding proceeds by alternately prompting the base model with two distinct system instructions $\mathcal{I}_R$ and $\mathcal{I}_S$ (see Appendix~\ref{app:prompts}):
\begin{align}
\hlorange{\z_R^{(\mathrm{t})}} &\sim \pi_\theta\!\left(\cdot \mid \mathcal{I}_R, \x, \hlblue{\z_S^{(\mathrm{t}-1)}}\right), \\
\hlblue{\z_S^{(\mathrm{t})}} &\sim \pi_\theta\!\left(\cdot \mid \mathcal{I}_S, \x, \hlorange{\z_R^{(\mathrm{t})}}, \hlblue{\z_S^{(\mathrm{t}-1)}}\right). \label{rc_summ}
\end{align}
$\mathcal{I}_R$ instructs the model to generate reasoning conditioned on the current cache, while $\mathcal{I}_S$ instructs the model to compress the current reasoning trace and previous summary into an updated summary that encodes high-level information about the strategies employed and conclusions reached in previous turns. After $\mathrm{T}$ turns, the final output is given by $\z := \z_R^{(\mathrm{T})}$. We denote this iterative process (see Figure~\ref{fig:main_fig}) as:
\begin{align}
(\hlorange{\z_R^{(\mathrm{1})}}, \hlblue{\z_S^{(\mathrm{1})}}, \dots, \hlblue{\z_S^{(\mathrm{T}-1)}}, \hlorange{\z_R^{(\mathrm{T})}}) 
	\sim \texttt{Alg}(\pi_\theta; \x).
\end{align}
\textbf{Extrapolation with \methodname.}
Because each step is allocated a fixed budget $H_R + H_S$, the total effective budget under \methodname{} is $\mathrm{T} \times (H_R \times H_S)$. Since $H_R \gg H_S$, we drop $H_S$ and approximate the budget as $\mathrm{T} \times H_R \overset{\mathrm{def}}{:=} H_\text{test}$. If performance improves in the regime $\mathrm{T'} \times H_R \gg H_{\mathrm{train}}$, we say that \methodname{} enables extrapolation.

\vspace{-0.2cm}
\subsection{Experimental Evaluation}
\label{subsec:rc_eval}
\vspace{-0.1cm}
\textbf{Experimental setup.} We now validate whether LLMs possess the ability to to utilize \methodname{} without additional training. We evaluate \methodname{} decoding with Qwen3-4B-Instruct-2507 and Qwen3-30B-A3B-Instruct-2507~\citep{qwen3technicalreport}, two hybrid LLMs capable of both complex reasoning and instruction-following (see Appendix~\ref{app:gptoss} for similar results from another model family). Using these LLMs as our base models, we run \methodname{} decoding for $\mathrm{T} = \text{12}$ turns with $H_S = \text{2048}$ and $H_R = \text{16k}$, giving us a total budget of $H_\text{test} = 192\text{k}$. This is far larger than both models' $H_\text{train}$, which we estimate to be about 16k (see Appendix~\ref{app:H_R_choice} for evidence justifying this). We use the November version of HMMT 2025 as our evaluation dataset (this competition was conducted after the base models were released), and generate 16 \methodname{} outputs per problem.

\textcolor{lightblue}{\textbf{Finding 1: \methodname{} enables extrapolation.}} We plot how accuracy evolves with the token budget $H_\text{test}$ in Figure~\ref{fig:rc_analysis_1} (left). We find that \methodname{} extrapolates reasoning far beyond $H_\text{train} = 16\text{k}$: 4B model accuracy increases by 17\% as the test token budget is scaled from 16k to 192k, while 30B model accuracy increases by 12\%. We also plot how actual token usage varies with reasoning token budget in Figure~\ref{fig:rc_analysis_1} (middle). We find that the cumulative number of tokens used scales linearly with the provided budget, which indicates that the model utilizes additional test time compute as provided and does not substantially shorten its responses at later iterations. Overall, our findings demonstrate that \methodname{} satisfies both desiderata outlined in Section~\ref{sec:reasoning_cache} and thus enables effective extrapolation.

\textcolor{lightblue}{\textbf{Finding 2: Summary-based abstractions are key to effective extrapolation.}} We examine the role of summarization–generation asymmetry by experimenting with iterative decoding methods that do not utilize summarization. Concretely, we remove the summary step and instead condition each iteration directly on the full response from the previous iteration $\z_R^{(\mathrm{t})}$, and prompt Qwen3-4B-Instruct-2507 ($H_R = 16\text{k}$) to either verify-then-correct (self-verify) or self-refine its solution (details in Appendix~\ref{app:iter_baselines}). Figure~\ref{fig:rc_analysis_1} (right) shows that \methodname{} consistently outperforms these baselines across all $H_\text{test}$ values, demonstrating that summary-conditioned generation provides benefits over other iterative methods. We attribute this to two factors: first, conditioning on summaries keeps context lengths bounded and in-distribution, whereas iterating on raw responses exceeds $H_{\text{train}}$ and induces distribution shift. Second, summarization serves to remove redundant “distractor” tokens~\citep{hong2025context, liu2023lostmiddlelanguagemodels} that obfuscate key findings and other important information, thereby yielding clearer guidance for subsequent reasoning.

\begin{figure*}[t]
    \centering
    \vspace{-0.2cm}
    \begin{subfigure}[b]{0.99\textwidth}
    \captionsetup{font=small,skip=2pt}
        \centering
        \includegraphics[width=0.99\textwidth]{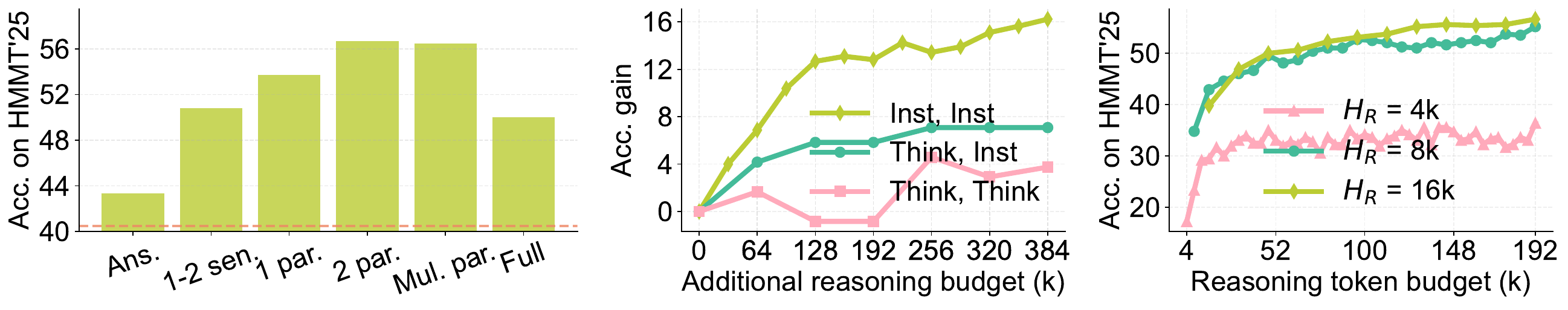}
        \caption*{}
    \end{subfigure}\hfill
    \vspace{-0.6cm}
    \caption{\footnotesize{\textbf{\textit{Left:} Accuracy at different levels of summary detail.} All accuracies measured at $H_\text{test} = 192k$; red dotted line indicates base model performance without \methodname{}. Performance degrades when summary detail is either too high or too low. \textbf{\textit{Middle:} Relative accuracy improvement (over standard autoregressive decoding) on HMMT 2025.} Replacing Qwen3-4B-Instruct-2507 with Qwen3-4B-Thinking-2507 for summarization (Think, Inst) reduces gains. Using Qwen3-4B-Thinking-2507 for both tasks (Think, Think) further reduces gains. \textbf{\textit{Right:} Accuracy vs. test-time token budget with various $H_R$.} Reducing $H_R$ from 16k to 8k leaves \methodname{} performance unchanged, whereas decreasing it further to 4k negatively impacts performance.}}
    \label{fig:rc_analysis_2}
    \vspace{-0.2cm}
\end{figure*}
\textcolor{lightblue}{\textbf{Finding 3: Summary detail level matters.}} Next, we study how much information summaries should retain. We vary the prompt $\mathcal{I}_S$ to produce summaries of differing detail, ranging from answer-only to multiple paragraphs (Figure~\ref{fig:summ_length_prompt}), and compare these to our default approach and to full-trace conditioning (self-refinement). Figure~\ref{fig:rc_analysis_2} (left) reports the accuracy of Qwen3-4B-Instruct-2507 at $H_{\text{test}} = 192\text{k}$ across various choices of $\mathcal{I}_S$. Performance degrades with very short summaries, improves as more detail is added, and peaks with $\geq$2-paragraph summaries; omitting summarization entirely degrades performance.

\begin{wrapfigure}{r}{0.35\textwidth}
    \centering
    \vspace{0cm}
    \includegraphics[width=0.33\textwidth]{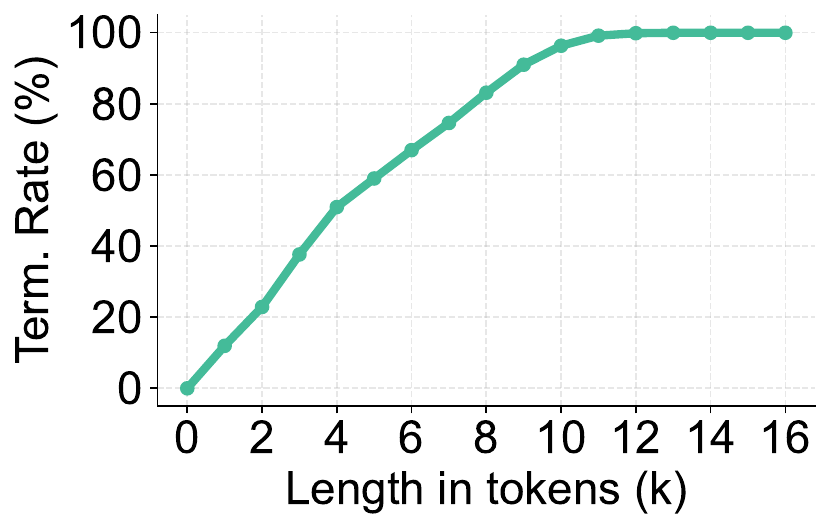}
    \vspace{-0.3cm}
    \caption{\footnotesize{\textbf{Termination rates of Qwen3-4B-Instruct-2507 with RC as a function of length.} Measured on HMMT 2025 and across all $\mathrm{T} = 12$ turns. Virtually all reasoning traces terminate within 16k tokens, suggesting that the model is not trained to reason beyond this. We consider a trace to have terminated after it generates \texttt{boxed\{\}}.}}
    \label{fig:term_plot}
    \vspace{-0.2cm}
\end{wrapfigure}
\textcolor{lightblue}{\textbf{Finding 4: Base models must be good instruction-followers for \methodname{} to be effective.}} We replace Qwen3-4B-Instruct-2507 with the specialist reasoning model Qwen3-4B-Thinking-2507, which excels at reasoning but possesses weaker instruction-following abilities and thus a less clear summarization-generation asymmetry. We evaluate using the reasoning model only for summary-conditioned generation (``Think, Inst'') and for both summary and summary-conditioned generation (``Think, Think'') ($H_R = 64\text{k}$). Figure~\ref{fig:rc_analysis_2} (middle) shows that ``Think, Inst'' and ``Think, Think'' both achieve smaller relative gains; qualitative inspection reveals that the reasoning model sometimes ignores summaries during generation and omits key details during summarization. However, note that \methodname{} still provides positive performance gains in all configurations, suggesting that sufficient asymmetry is present for training to potentially be effective. Indeed, \citet{qednano2026} demonstrate that \methodname{} training substantially improves the proof-writing capabilities of Qwen3-4B-Thinking-2507, confirming that sufficient summarization-generation asymmetry exists even in specialist reasoning models for the method to be effective.

\textcolor{lightblue}{\textbf{Finding 5: Reducing $H_R$ by too much degrades performance.}} By default, we set $H_R = H_\text{train} = 16\text{k}$, and since LLMs rarely generate traces longer than $H_\text{train}$, we only consider decreasing it. Figure~\ref{fig:rc_analysis_2} (right) shows that reducing $H_R$ from 16k to 8k has minimal impact despite the fraction of incomplete traces increasing from 0\% to 20\% (see Figure~\ref{fig:term_plot}). Reducing $H_R$ further to 4k causes nearly 50\% of traces to terminate early, this time substantially degrading performance. This implies that while $H_R$ can be decreased, it must still be large enough for redundancy to emerge. When $H_R$ is set too small, the resulting summaries capture only shallow progress that provides insufficient signal for continuation, which encourages the model to restart reasoning from scratch instead.

\vspace{-0.2cm}
\subsection{Analysis of Summary-Conditioned Generations}
\label{subsec:rc_traces} 
\vspace{-0.1cm}

\begin{wrapfigure}{r}{0.35\textwidth}
    \centering
    \vspace{-1.2cm}
    \includegraphics[width=0.33\textwidth]{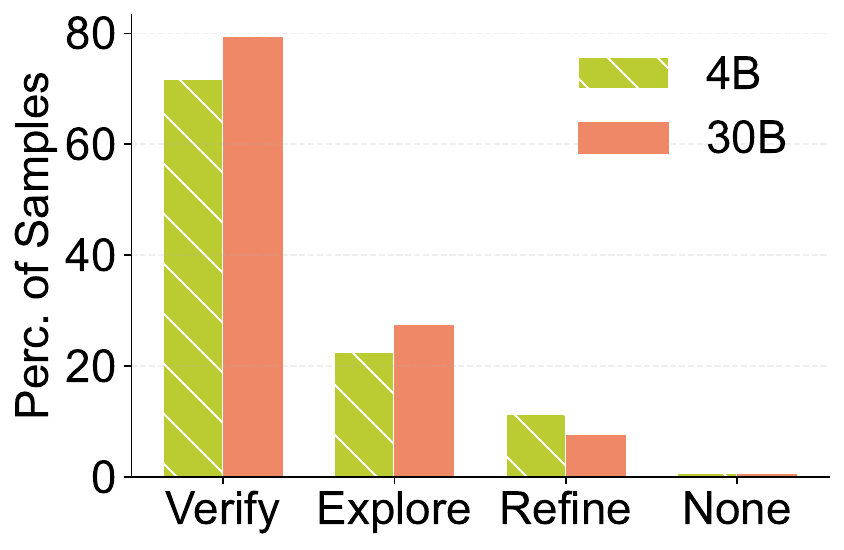}
    \vspace{-0.3cm}
    \caption{\footnotesize{\textbf{Percentage of \methodname{} reasoning traces that employ various reasoning strategies.} \textit{Verification} of previous reasoning is the most common strategy, followed by \textit{exploration} of new strategies.}}
    \label{fig:category_comp}
    \vspace{-0.2cm}
\end{wrapfigure}
We analyze the content of summary-conditioned generations produced by \methodname{} and find that they commonly exhibit three high-level strategies: \textbf{(1)} \emph{verification}, where the model generates reasoning to explicitly verify intermediate or final results stated in the summary; \textbf{(2)} \emph{exploration}, where the model deliberately pursues a different strategy from that used in the summary; and \textbf{(3)} \emph{refinement}, where the model acknowledges the summary and repeats the same strategy without attempting verification or exploring alternatives. To quantify the prevalence of these behaviors, we extract summaries and their subsequent reasoning traces and pass them to an LLM-based annotator (Figure~\ref{fig:assess_prompt}), which assigns each sample to one of the three categories above, as well as a \emph{none} category when no references to the summary are made and reasoning restarts from scratch. Figure~\ref{fig:category_comp} shows that the model relies heavily on summaries to guide subsequent generations, with very few samples classified as \emph{none}. The most common strategy is \emph{verification}, although a substantial fraction of samples also exhibit \emph{exploration} and \emph{refinement}.

\begin{AIbox}{Key Takeaways: Inference-only experiments with \methodname{}}
\begin{itemize}[leftmargin=0.7em]
    \setlength\itemsep{0em}
    \item  \methodname{} enables extrapolation even without \methodname{}-specific training. It exploits the \emph{summarization-generation asymmetry} and is thus most effective when paired with instruction-following models.
    
\end{itemize}
\end{AIbox}
\vspace{-0.2cm}
\section{Training Models to Extrapolate with \methodname{} Decoding}
\label{sec:rc_training}
\vspace{-0.2cm}

\begin{figure}[t]
    \centering
    \vspace{-0.2cm}
    \begin{subfigure}[b]{0.99\textwidth}
    \captionsetup{font=small,skip=2pt}
        \centering
        \includegraphics[width=0.99\textwidth]{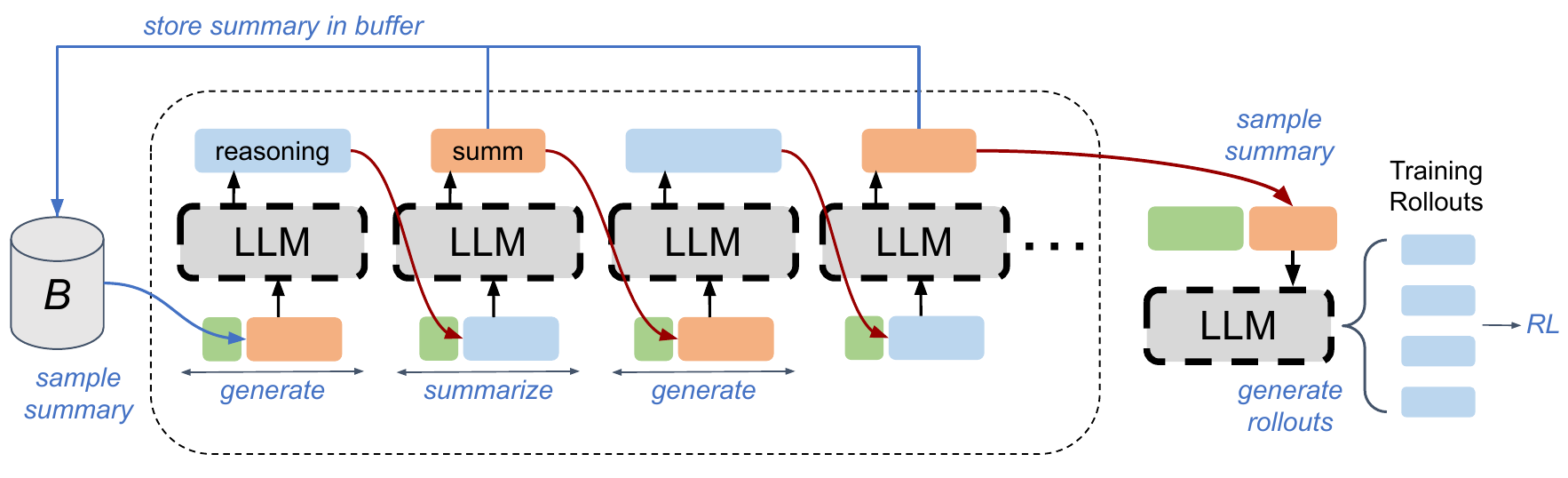}
        \caption*{}
    \end{subfigure}\hfill
    \vspace{-0.6cm}    
    \caption{\footnotesize{\textbf{Illustration of \methodname{} rollout generation for training with a replay buffer $\mathcal{B}$.} We sample an input and a summary from $\mathcal{B}$, and use this as the starting point to run $\mathrm{T}_\text{train}$ steps of \methodname{} decoding. New summaries generated through this process are stored in $\mathcal{B}$, replacing older summaries corresponding to the same input problem. We then sample from the $\mathrm{T}_\text{train}$ newly generated summaries, and condition on these to generate training rollouts that are optimized via Equation~\ref{eq:train_obj}.}}
    \label{fig:train_diag}
    \vspace{-0.2cm}
\end{figure}

Having established the design of \texttt{Alg}, we now describe our method for training models to use it. Our analysis in Section~\ref{subsec:rc_eval} shows that \methodname{} extrapolation depends on the model’s ability to iteratively reason from and improve upon summaries of past iterations. Accordingly, our training objective is to strengthen \emph{summary-conditioned generation}: given a problem and a summary, we should train our model to generate improved reasoning that is more likely to yield a correct answer. The iterative structure of \methodname{} makes this objective amenable to standard outcome-reward RL as we can run \methodname{} for multiple iterations and apply gradient updates independently at each turn. By setting $H_R$ to a ``typical'' value for the base model, such that most responses end with a final answer, we can assign outcome-based rewards at each step and avoid credit assignment across iterations. While this approach is indeed myopic, we note that the iterative nature of \methodname{} decoding ensures that this does not reinforce the resulting bias of premature termination.

Formally, we run \methodname\ for $\mathrm{T}_\text{train}$ turns for each problem $\x$ in a training batch. We collect the summaries generated from this $\z_S := (\z_S^{(1)}, \dots, \z_S^{(\mathrm{T}_\mathrm{train})})$ and uniformly sample $N_{\text{summ}} \leq \mathrm{T}_{\mathrm{train}}$ unique summaries per problem. We then generate $K$ reasoning traces conditioned on each sampled summary, assign outcome rewards and compute advantages over these samples:
\vspace{-0.2cm}
\begin{align}
    \max_{\pi_\theta}~~~ &\mathbb{E}_{\x, \y \sim \mathcal{D}_{\text{train}},~ t \sim U[1, \mathrm{T}_{\mathrm{train}}]} \left[ \mathbb{E}_{\z^\prime \sim \pi_{\theta}(\cdot|\mathbf{x}, \z_S^{(t)})} \left[r(\y, \z^\prime)\right]\right],\nonumber \\ 
    &\text{where}\quad(\z_S^{(1)}, \dots, \z_S^{(\mathrm{T}_\text{train})})
    \sim \texttt{Alg}(\pi_\theta; \x),\quad
    \text{s.t.}~~  |\z^\prime| \leq H_R.
    \label{eq:train_obj}
\end{align}
This design is effective because, under the problem structure targeted by \methodname{}, optimizing each iteration locally is aligned with optimizing the full reasoning trajectory. By training each step $\mathrm{t}$ to produce a correct answer, we implicitly encourage the model to generate reasoning traces and summaries that contain information, such as intermediate results or partial analyses, that is useful for subsequent steps. Because training rollouts are conditioned on these summaries, the model is also explicitly trained to exploit this accumulated information when generating later responses, increasing the likelihood of future correctness. Under the assumption that successive summarization and summary-conditioned generation steps make roughly monotonic progress toward a solution, optimizing per-step correctness is sufficient to improve trajectory-level performance, despite only explicitly optimizing individual steps. We discuss limitations of this assumption and evaluate alternative training approaches in Appendix~\ref{app:limitations}.

\textbf{Training with a summary replay buffer (off-policy reinforcement learning).} The iterative decoding structure of \methodname{} naturally enables learning from off-policy summaries because summaries serve as conditioning inputs rather than optimization targets. Learning from off-policy summaries provides two benefits. First, it enables training on later reasoning turns without the need for us to generate long on-policy decoded trajectories, which is useful because summaries from later turns may qualitatively differ from earlier ones. Second, it increases the coverage of summaries the model encounters during training, which in turn increases the robustness of the policy to test time shifts in the summary distribution. In fact, improving state coverage via a replay buffer is one of the fundamental principles in off-policy RL~\citep{fu2019diagnosing}, which using past summaries enables us to implement.

We therefore incorporate a \emph{summary replay buffer} into training; see Figure~\ref{fig:train_diag}. During the first training epoch, we follow the same on-policy procedure to optimize Equation \ref{eq:train_obj}, but also store all generated summaries, and their corresponding problems, in the replay buffer $\mathcal{B}$. From the second epoch onward, we sample problems and summaries from $\mathcal{B}$ and condition \methodname{} rollouts on them instead of generating fresh summaries, thereby extending the maximum effective training horizon by $\mathrm{T}_\mathrm{train}$ steps per epoch.
\vspace{-0.2cm}
\section{Experimental Evaluation: Training with \methodname{}}
\label{sec:results}
\vspace{-0.2cm}
The goal of our experiments is to demonstrate the effectiveness of training with \methodname{} to improve extrapolation. To this end, we evaluate our approach on several benchmarks, compare it with related methods, and also conduct several ablation experiments to isolate the effects of our different design choices. 

\textbf{Training details.} We post-train a Qwen3-4B-Instruct-2507 model to utilize \methodname{} and refer to the trained model as \texttt{RCT-4B}. We set $K = 8$, $N_\text{summ} = 2$ and $\mathrm{T}_{\text{train}} = 3$. We conduct training in two stages: in Stage I, we train without the summary replay buffer, focusing on optimizing early turns, including the initial turn (without any summary context). Training problems for Stage I are subsampled from the AceReason-Math dataset~\citep{chen2025acereason}, resulting in a dataset of about 5.7k problems. For Stage II, we enable the summary replay buffer, and construct a new training set by injecting a small number of difficult problems from DAPO~\citep{yu2025dapoopensourcellmreinforcement} into our Stage I dataset as part of our training curriculum. See Appendix~\ref{app:data_construction} for details.

\textbf{Benchmarks and evaluation protocols.} We evaluate \texttt{RCT-4B} on three math reasoning benchmarks: \textbf{AIME 2025}, \textbf{HMMT 2025} (November version), and \textbf{IMO-AnswerBench}~\citep{luong2025robustmathematicalreasoning}, as well as one scientific reasoning benchmark, \textbf{FrontierScience} (Olympiad)~\citep{openai2025frontierscience}, which contains expert-written problems in physics, chemistry, and biology. Since our training data exclusively consists of mathematical reasoning problems, FrontierScience serves to assess whether learned extrapolation behavior generalizes to an unseen domain. We evaluate AIME and HMMT using \texttt{math-verify}\footnote{\url{https://github.com/huggingface/Math-Verify}} for equivalence checking, and follow the official LLM judge-based evaluation protocols for IMO-AnswerBench\footnote{In a previous version of this paper, we reported IMO-AnswerBench results evaluated using \texttt{math-verify}. We subsequently discovered that this approach underestimates accuracy, as \texttt{math-verify} fails to parse certain valid answer formats on this benchmark, leading to correct responses being marked as incorrect. We therefore report results using the official LLM judge-based evaluation protocol instead, replacing Gemini-2.5-Pro with Gemini-3-Flash.} and FrontierScience. We selected the latter three benchmarks for their low contamination risk: all three were released after our training datasets and our base model, with HMMT 2025 (Nov) and FrontierScience consisting entirely of new problems and IMO-AnswerBench consisting of extensively rewritten past Olympiad problems.

\textbf{Baselines and comparisons.} We compare against three categories of approaches. The first consists of \textbf{autoregressive decoding} methods using open-source 4B models. These include Qwen3-4B-Instruct-2507 (our base model), Qwen3-4B-Thinking-2507, Polaris-4B~\citep{Polaris2025}, a Qwen3-4B-based model trained for extrapolation by expanding the output context using YaRN~\citep{peng2023yarnefficientcontextwindow}, and a version of Qwen3-4B-Instruct-2507 trained with standard GRPO at $H_\text{train} = 32\text{k}$, the maximum output length we could reliably use for RL due to practical constraints. The second category of comparisons consists of other \textbf{iterative decoding} approaches that condition directly on raw past reasoning rather than on summaries. We evaluate base and trained (see Appendix~\ref{app:iter_baselines}) versions of two such methods: self-refinement and self-verification,  selecting $H_R$ and $H_\text{test}$ to be the same as for our \methodname{} experiments. We also evaluate two other iterative decoding approaches inspired by recent work~\citep{muennighoff2025s1simpletesttimescaling, aghajohari2025markovianthinkerarchitectureagnosticlinear} in Appendix~\ref{app:additional_baselines}. The third category of comparisons consists of approaches that do not train with \methodname{} but still use \methodname{} at inference time. We compare \texttt{RCT-4B} against inference-only use of \methodname{} by the base model (as in Section~\ref{subsec:rc_eval}) and by the base model post-trained with standard RL. This comparison isolates the contribution of training with \methodname{}, rather than applying it solely at test time. See Appendix~\ref{app:efficiency} for a discussion on the computational efficiency of \methodname{}.

\vspace{-0.2cm}
\subsection{Benchmark Results}\label{sec:train_bench}
\vspace{-0.2cm}

\begin{figure}[t]
    \centering
    \vspace{-0.2cm}
    \begin{subfigure}[b]{0.99\textwidth}
     \captionsetup{font=small,skip=2pt}
        \centering
        \includegraphics[width=0.99\textwidth]{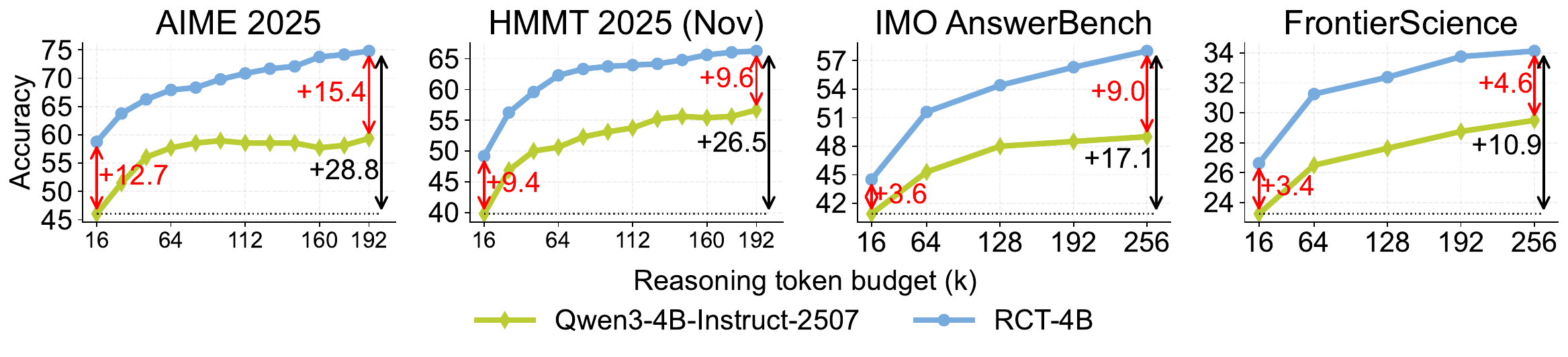}
        \caption*{}
    \end{subfigure}\hfill
    \vspace{-0.6cm}
    \caption{\footnotesize{\textbf{Accuracy on various reasoning benchmarks as a function of token budget for Qwen3-4B-Instruct-2507 and \texttt{RCT-4B} with \methodname{} decoding.} \texttt{RCT-4B} improves performance across all four reasoning benchmarks as reasoning token budget $H_\text{test}$ is increased beyond $H_\text{train}$. This improvement is larger than that attained by using the base model.}}
    \label{fig:4b_main_results}
    \vspace{-0.2cm}
\end{figure}

\begin{table}[h]
\centering
\footnotesize
\begin{tabular}{l | c c c c}
\toprule
& \textbf{AIME 2025} & \textbf{HMMT 2025 (Nov)} & \textbf{IMO-AnswerBench} & \textbf{FrontierScience} \\
\midrule
Qwen3-4B-Instruct-2507 [16k] & 46.0 & 39.8 & 40.9 & 23.3 \\
Qwen3-4B-Instruct-2507 (RL, 32k) [32k] & 54.8 & 48.3 & 42.1 & 21.5 \\
Polaris-4B~\citep{Polaris2025} [90k] & 79.4 & 60.2 & 47.7 & 23.6 \\
Qwen3-4B-Thinking-2507 [81k] & \textbf{81.3} & 62.5 & 53.8 & 25.7 \\
\midrule
Self-Refine (base) & 53.8 & 50.0 & 45.8 & 27.8 \\
Self-Verify (base) & 48.9 & 46.7 & 43.9 & 29.7 \\
Self-Refine (trained) & 60.4 & 61.3 & 52.2 & 33.5 \\
Self-Verify (trained) & 61.2 & 62.1 & 52.3 & 31.9 \\
\midrule
Qwen3-4B-Instruct-2507 + \methodname{} & 59.4 & 56.7 & 46.3 & 29.5 \\
Qwen3-4B-Instruct-2507 (RL, 32k) + \methodname{} & 66.0 & 60.2 & 53.4 & 33.5 \\
\textbf{\texttt{RCT-4B} + \methodname{}} \textcolor{red}{\textbf{\textit{(Ours)}}} & 74.9 & \textbf{66.3} & \textbf{58.0} & \textbf{34.1} \\
\bottomrule
\end{tabular}
\vspace{-0.2cm}
\caption{\footnotesize{\textbf{Evaluation results.} The top section of the table reports results of various reasoning models run with standard autoregressive generation; $H_\text{test}$ given in square brackets. The middle section reports various iterative decoding baselines, run for 12 (AIME, HMMT) or 16 (IMO-AnswerBench, FrontierScience) turns (equivalent to a 192k or 256k reasoning token budget) with $H_R = 16\text{k}$. The final section reports results for \methodname{}-based results, which were run with the same token budgets as the iterative decoding baselines.
\texttt{RCT-4B} outperforms all baseline methods on 3 out of 4 benchmarks.}}
\label{tab:4b_main_tab}
\vspace{-0.1cm}
\end{table}

Our results are shown in Figure~\ref{fig:4b_main_results} and Table~\ref{tab:4b_main_tab}. Across all benchmarks and token budgets, \texttt{RCT-4B} outperforms the base model using \methodname{}, with the performance gap widening as the token budget increases. This indicates that training enables more effective extrapolation rather than merely improving short-horizon performance. Notably, the model also improves on FrontierScience despite being trained exclusively on mathematics problems, suggesting that \methodname{} training develops domain-general extrapolation capabilities.

\textcolor{lightblue}{\textbf{Finding 1: \methodname{} compares favorably against other strong 4B-sized models.}} We compare \texttt{RCT-4B} against strong 4B reasoning-specialized models that utilize autoregressive decoding. While these models are explicitly trained to exploit large token budgets, \texttt{RCT-4B} outperforms all autoregressive approaches on the three benchmarks with lowest contamination risk: HMMT 2025 (released in Nov '25), IMO-AnswerBench (released in Nov '25), and FrontierScience (released in Dec '25). In fact, we find that \texttt{RCT-4B} even achieves competitive results against much larger reasoning models: see Figure~\ref{fig:imo_extra_compare}. Notably, while standard RL training improves upon the base model on the mathematics benchmarks, this standard RL model \textbf{(1)} remains substantially weaker than specialized reasoning models, and \textbf{(2)} achieves no gains on the out-of-domain FrontierScience benchmark despite training on the same data as \texttt{RCT-4B}. This demonstrates that \methodname{} training develops more generalizable problem-solving strategies that standard RL does not.

\textcolor{lightblue}{\textbf{Finding 2: \methodname{} training yields better iterative reasoning than other iterative training methods.}} \texttt{RCT-4B} substantially outperforms all iterative decoding methods we compare against. Training models to perform self-verification or self-refinement using an approach analogous to \methodname{} training (Appendix~\ref{app:iter_baselines}) leads to large improvements over their respective base models. However, these baselines remain significantly weaker than \texttt{RCT-4B} on mathematical reasoning tasks, highlighting the benefit of exploiting the summarization–generation asymmetry during training. Interestingly, nearly all iterative decoding methods we evaluate, including untrained variants, outperform autoregressive decoding on FrontierScience. This suggests that iterative decoding may generalize better to out-of-domain input problems compared to standard autoregressive generation for the base models we consider in our experiments. 

\vspace{-0.25cm}
\subsection{Evaluating \methodname\ on Hard Problems}\label{sec:train_diff}
\vspace{-0.1cm}

Our results thus far show that extrapolation with \texttt{RCT-4B} yields consistent improvements on standard benchmarks. However, these gains could have arisen either from solving harder problems with additional test-time compute or from sharpening performance on problems that are already partially solvable within $H_R$. To distinguish between these effects, we evaluate models on a set of adversarially curated problems from Omni-MATH~\citep{gao2024omnimathuniversalolympiadlevel}, following the protocol of \citet{qu2025learning}. These problems are selected such that the base Qwen3-4B-Instruct-2507 model fails to produce any correct solution across $N=256$ independent attempts, even when given sufficient reasoning budget. We consider such problems to be difficult because they are unlikely to be solvable simply by scaling parallel compute~\citep{cobbe2021trainingverifierssolvemath, wang2023selfconsistencyimproveschainthought, wu2025betterinstructionfollowingminimumbayes}.

\begin{figure}[t]
    \centering
    \vspace{-0.2cm}
    \begin{subfigure}[b]{0.99\textwidth}
     \captionsetup{font=small,skip=2pt}
        \centering
        \includegraphics[width=0.99\textwidth]{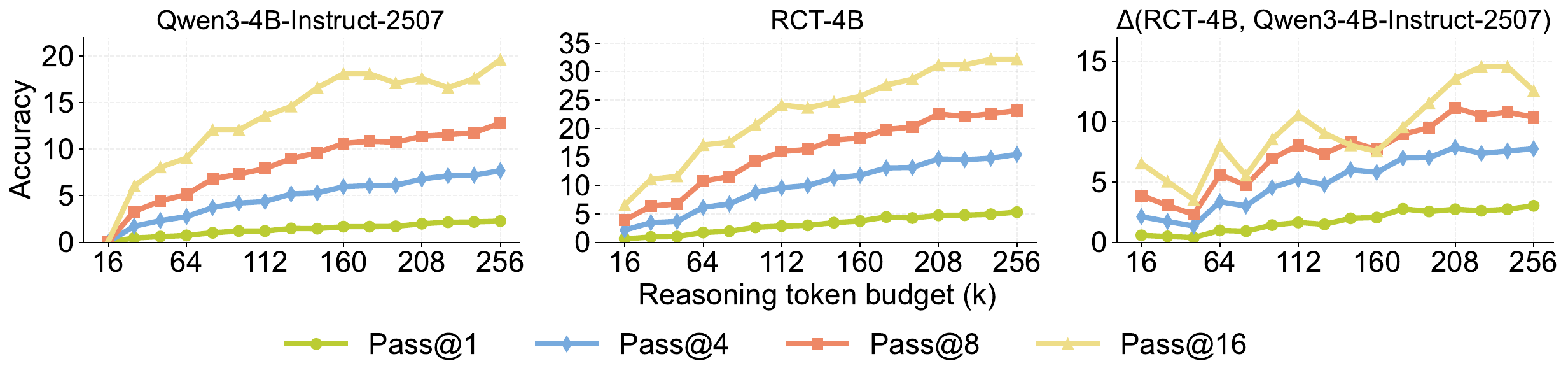}
        \caption*{}
    \end{subfigure}\hfill
    \vspace{-0.6cm}
    \caption{\footnotesize{\textbf{Pass@$k$ accuracy vs. token budget on a hard subset of problems sampled from \cite{qu2025learning} for Qwen3-4B-Instruct-2507 and \texttt{RCT-4B}}. The left panel shows base model performance, the middle panel shows trained model performance, and the right panel shows the performance gap between the two. \texttt{RCT-4B} achieves substantially higher pass@$k$ rates than the base model across all values of $k$, with the performance gap increasing as reasoning token budget grows.}}
    \label{fig:omni_plots}
    \vspace{-0.2cm}
\end{figure}

\textbf{Evaluation results.} We evaluate both the base model and \texttt{RCT-4B} on this dataset using \methodname{} decoding and report pass@$k$ in Figure~\ref{fig:omni_plots}. While both models improve with increased reasoning token budgets, the gains for \texttt{RCT-4B} are substantially larger. At a budget of 256k tokens, the base model achieves a pass@$16$ of 20\%, whereas \texttt{RCT-4B} reaches nearly 35\%. Moreover, the performance gap between the trained and base models widens as the token budget increases, indicating that training improves the model’s ability to utilize \methodname{} for in-context exploration. Overall, these results demonstrate that \texttt{RCT-4B} can solve difficult problems that the base model cannot by effectively extrapolating its reasoning at test time. For additional results directly comparing \methodname{} against parallel compute methods (majority vote), see Appendix~\ref{app:maj_vote}.

\vspace{-0.25cm}
\subsection{Ablation Studies}\label{sec:train_ablate}
\vspace{-0.1cm}

\begin{figure}[t]
    \centering
    \vspace{-0.2cm}
    \begin{subfigure}[b]{0.99\textwidth}
     \captionsetup{font=small,skip=2pt}
        \centering
        \includegraphics[width=0.99\textwidth]{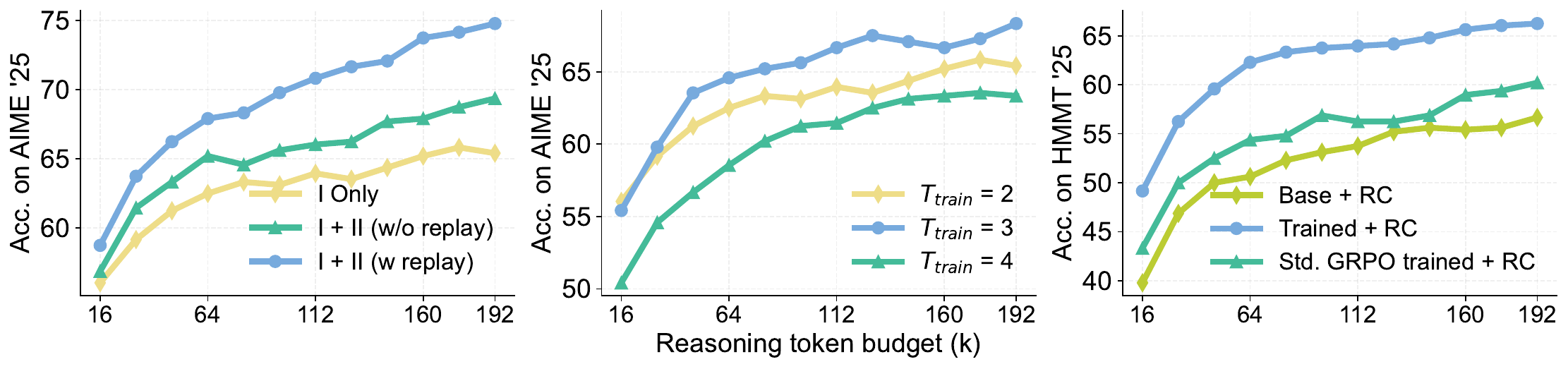}
        \caption*{}
    \end{subfigure}\hfill
    \vspace{-0.6cm}
\caption{\footnotesize{\textbf{\textit{Left:} Ablation study on stagewise training configurations.} Performance improves with Stage II training, with the summary replay buffer providing additional gains that increase with reasoning budget. \textbf{\textit{Middle:} Ablation study on $\mathrm{T}_\text{train}$ values during Stage I.} $\mathrm{T}_\text{train} = 3$ yields optimal performance across budgets by balancing early-turn training with summary-conditioned reasoning exposure. \textbf{\textit{Right:} Comparison of training methods.} \methodname{}-specific training substantially outperforms standard GRPO training, demonstrating that extrapolation requires explicit training beyond simply improving reasoning capabilities.}}
    \label{fig:ablate}
    \vspace{-0.2cm}
\end{figure}

\textcolor{lightblue}{\textbf{The effect of the summary replay buffer.}} We  ablate our training procedure by comparing performance on AIME 2025 after \textbf{(i)} Stage I training only, \textbf{(ii)} Stage II training without using the summary replay buffer, and \textbf{(iii)} Stage II training with the replay buffer enabled; see Figure~\ref{fig:ablate} (left). Stage I training alone yields substantial gains over the base model, while Stage II training provides additional improvements. These gains are modest without the replay buffer but significantly larger when it is used, particularly at higher reasoning budgets. At 16k tokens, Stage II with replay buffer improves accuracy by 2.7\% over Stage I, while at 192k tokens this gap grows to 9.4\%. In contrast, Stage II without replay buffer yields only a $\sim$4\% improvement at 192k tokens, thus demonstrating the effectiveness of our replay buffer.

\textcolor{lightblue}{\textbf{The effect of the number of training turns.}} Next, we vary the number of training turns $\mathrm{T}_\text{train}$ during Stage I, evaluating $\mathrm{T}_\text{train} \in \{2, 3, 4\}$; see Figure~\ref{fig:ablate} (middle). All settings improve extrapolation, with $\mathrm{T}_\text{train} = 3$ performing best, followed by $\mathrm{T}_\text{train} = 2$. We believe this arises from a trade-off in gradient signal allocation: when  $\mathrm{T}_\text{train}$ is too large, the model receives insufficient signal on early turns (including the initial turn), while the converse is true when $\mathrm{T}_\text{train}$ is too low.
These results imply that we must train the model evenly across all early turns in order for good summary-conditioned reasoning to emerge. 


\textcolor{lightblue}{\textbf{The importance of training with \methodname{}.}} Finally, we isolate the effect of \methodname{} training by comparing against a model trained with standard outcome-reward RL (GRPO). We train Qwen3-4B-Instruct-2507 with RL for the same number of steps and following the same two-stage training curriculum while increasing the training budget $H_\text{train}$ from 16k to 32k tokens (the maximum output length we could reliably use for RL due to practical constraints). We then conduct evaluation using \methodname{} decoding; see Figure~\ref{fig:ablate} (right) and Table~\ref{tab:4b_main_tab} (bottom) for our results. While RL training yields modest improvements over the base model, it falls far short of training with \methodname{}.  This demonstrates that effective extrapolation via \methodname{} decoding does not emerge from standard RL alone and must be trained through \methodname{}’s structured, multi-turn objective.
\newpage

\vspace{-0.2cm}
\subsection{Incorporating \methodname{} into test-time scaffolds}\label{sec:scaffolds}
\vspace{-0.1cm}

\begin{wraptable}{r}{0.35\textwidth}
\centering
\footnotesize
\begin{minipage}{\linewidth}
\centering
\resizebox{0.99\linewidth}{!}{\begin{tabular}{c|cc}
\toprule
 & RSA & DSM Agent \\
\midrule
Base & 66.3 & 57.5 \\
RL & 64.8 & 61.3 \\
\texttt{RCT-4B} (no \methodname{}) & 70.2 & 65.8 \\
\midrule
\texttt{RCT-4B} + \methodname{} \textcolor{red}{\textit{\textbf{(Ours)}}} & \textbf{75.4} & \textbf{74.6} \\
\bottomrule
\end{tabular}}
\vspace{-0.2cm}
\caption{\footnotesize\textbf{\texttt{RCT-4B} yields additional gains when incorporated into test-time scaffolds.} These gains are higher than with the base or standard RL-trained model. Using \methodname{} decoding within the scaffold yields further gains.}
\label{tab:scaffolds}
\end{minipage}
\vspace{-0.3cm}
\end{wraptable}
Our experiments thus far show that \methodname{} training enables more effective extrapolation by improving summary-conditioned generation. We can view this as a specific instance of \emph{self-guided reasoning}: the model conditions on self-generated abstractions~\citep{qu2025rladtrainingllmsdiscover, yang2026intselfproposedinterventionsenable} (in our case, summaries of prior reasoning) to guide downstream reasoning. This raises a natural question: does \methodname{} training teach a generalizable skill for using self-generated guidance, or are improvements narrowly tied to reasoning from summaries? To answer this, we evaluate how \methodname{} training transfers to other self-guidance settings by incorporating \texttt{RCT-4B} and \methodname{} decoding into two test-time scaffolds: Recursive Self-Aggregation (RSA)~\cite{venkatraman2025recursiveselfaggregationunlocksdeep} and the DeepseekMath (DSM) Agent~\citep{shao2025deepseekmathv2selfverifiablemathematicalreasoning}. RSA generates parallel reasoning traces and iteratively aggregates them, while the DSM Agent iteratively performs self-verification and self-refinement over an initial pool of solutions; see Appendix~\ref{app:scaffold_details}. In both cases, the model is used not only to generate reasoning traces from scratch, but also to reason conditioned on self-generated context produced via the scaffold.

\textbf{Evaluation results.} We report results on HMMT 2025 (Nov) in Table~\ref{tab:scaffolds}, which shows that \texttt{RCT-4B} leverages both scaffolds far more effectively than the base or RL-trained Qwen3-4B-Instruct-2507 models, even without \methodname{} decoding. This suggests that \methodname{} training imparts a generalizable capability: reasoning effectively from self-generated abstractions beyond summaries, including aggregated past traces (RSA) and self-generated feedback (DSM Agent). Replacing standard autoregressive decoding with \methodname{} for all solution-generation steps within these scaffolds yields additional gains, which we attribute to improved reasoning accuracy throughout the scaffolded process. Overall, these results indicate that \methodname{} training develops abstraction-conditioned reasoning abilities, highlighting an important direction for future work.

\begin{AIbox}{Key Takeaways: Training with \methodname{} improves extrapolation across domains}
\begin{itemize}[leftmargin=0.7em]
    \setlength\itemsep{0em}
    \item \texttt{RCT-4B} outperforms iterative decoding and autoregressive decoding baselines.
    \item \methodname{} enables models to solve difficult problems that cannot be solved by scaling parallel compute. \item \methodname{} training yields models that are better at summary- and abstraction- conditioned reasoning in general, and can therefore effectively leverage scaffolds to further scale test-time compute.
\end{itemize}
\end{AIbox}
\vspace{-0.2cm}
\section{Conclusion and Perspectives on Future Work}
\vspace{-0.1cm}
\label{sec:conclusion}

In this work, we demonstrate how LLMs can be trained to continually improve their reasoning across long horizons. Our method, \methodname{}, replaces autoregressive decoding with an iterative decoding algorithm that alternates between summarization and summary-conditioned reasoning, and trains models via outcome-reward RL to leverage this algorithm more effectively. We show that using \methodname{} allows us to overcome a fundamental limitation of autoregressive generation and standard RL: the inability to extrapolate reasoning beyond training rollout lengths. Empirically, we demonstrate that our \methodname{}-trained model achieves substantial performance gains on challenging mathematical and scientific benchmarks by extrapolating reasoning to much longer horizons than it was trained for. Furthermore, we show that extrapolation via \methodname{} enables the model to solve difficult problems that it cannot solve within its training budget, and that the trained model can leverage existing test-time scaffolds to further scale inference compute. Ultimately, we believe that \methodname{} represents an important step toward training models that can engage in the systematic, long-horizon reasoning required to solve the world's most difficult problems.

\textbf{Future work.} Future work should address the limitations of \methodname{} (see Appendix~\ref{app:limitations}). We identify three main directions. First, improving the training objective beyond myopic rewards: our current approach assigns rewards based on individual trace correctness, which discourages multi-turn strategies in which early turns perform exploratory, low-reward procedures that are only exploited in later turns. As we discussed in Section~\ref{subsec:rc_method}, we do not expect this to be a problem when  each iteration of summarization still makes useful progress, but training with a non-myopic extension of \methodname{} could potentially yield significant improvements on harder problems. Second, explicitly training for summary generation: our current approach only trains the model for summary-conditioned generation. Our experiments in Section~\ref{subsec:rc_eval}, however, show that altering  summaries can significantly affect the performance of \methodname{}, suggesting that directly optimizing summary generation could also be beneficial. This direction is also closely related to prior work on training models to produce useful abstractions for guiding reasoning~\citep{qu2025rladtrainingllmsdiscover, yang2026intselfproposedinterventionsenable} or in training aggregation mechanisms~\citep{venkatraman2025recursiveselfaggregationunlocksdeep}, with both lines of work demonstrating tangible performance gains. Third, adapting \methodname{} to tasks without final-answer based rewards: while our current work focuses on settings with closed-form, verifiable outcomes, these tasks represent only a minority of the reasoning challenges we ultimately care about. Broadening \methodname{} to operate in open-ended domains, such as proof generation, is therefore an exciting and potentially impactful direction for future research.

\vspace{-0.2cm}
\section*{Acknowledgements}
\vspace{-0.1cm}
\label{sec:ack}

We would like to thank Matthew Yang, Haoran Li, Lewis Tunstall, Jasper Dekoninck, and the other members of the CMU AIRe lab for discussions and feedback. We would also like to thank the CMU FLAME center for providing the compute resources on the Orchard cluster that supported almost all of our big experiments. Additionally, we are grateful for the Delta AI cluster at NCSA and the TRC program of Google Cloud for additional computational resources. This work was supported by the Office of Naval Research under ONR N0014-24-2206 and a Schmidt Sciences AI2050 Early Career Fellowship. AS is supported by a JP Morgan AI PhD fellowship and YQ is supported by an Amazon AI PhD fellowship.

\bibliography{main}


\newpage

\appendix
\part*{Appendices}

\section{Additional Results vs. Larger Reasoning Models}\label{app:larger_models}

\begin{figure}[htbp]
    \centering
    \begin{subfigure}[b]{0.99\textwidth}
     \captionsetup{font=small,skip=2pt}
        \centering
        \includegraphics[width=0.99\textwidth]{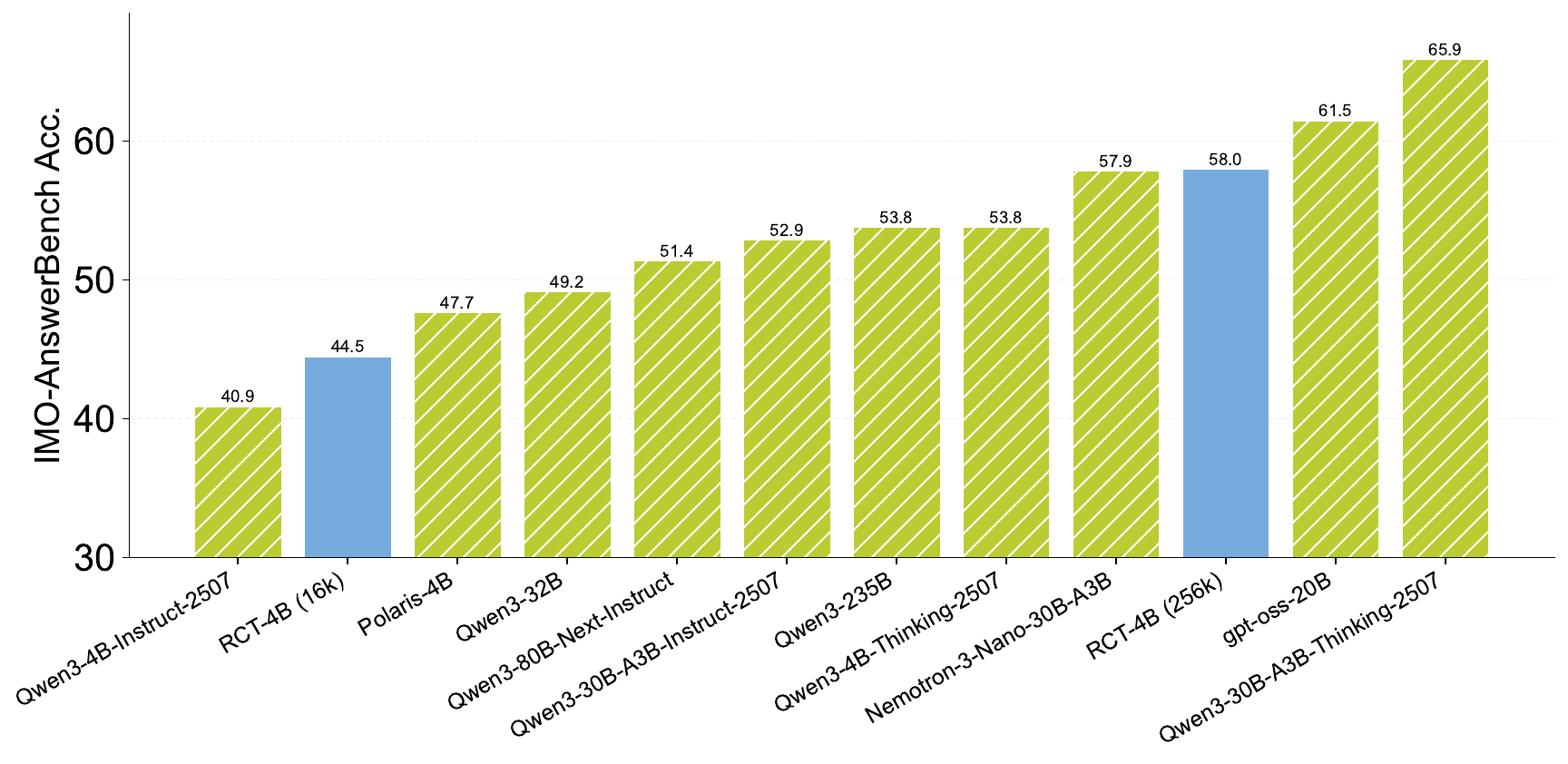}
        \caption*{}
    \end{subfigure}\hfill
    \vspace{-0.8cm}
\caption{\footnotesize{\textbf{Comparison of \texttt{RCT-4B} and a selection of other reasoning models on IMO-AnswerBench.} Combining \methodname{} training and decoding enables our 4B model to outcompete many larger and newer models. We set inference hyperparameters ($t$, $p$, $H_\text{test}$) based on the recommended values provided on each model's Hugging Face page.}}
    \label{fig:imo_extra_compare}
    \vspace{-0.2cm}
\end{figure}

\section{Comparison with Majority Voting}\label{app:maj_vote}

\begin{figure}[htbp]
    \centering
    \includegraphics[width=0.6\textwidth]{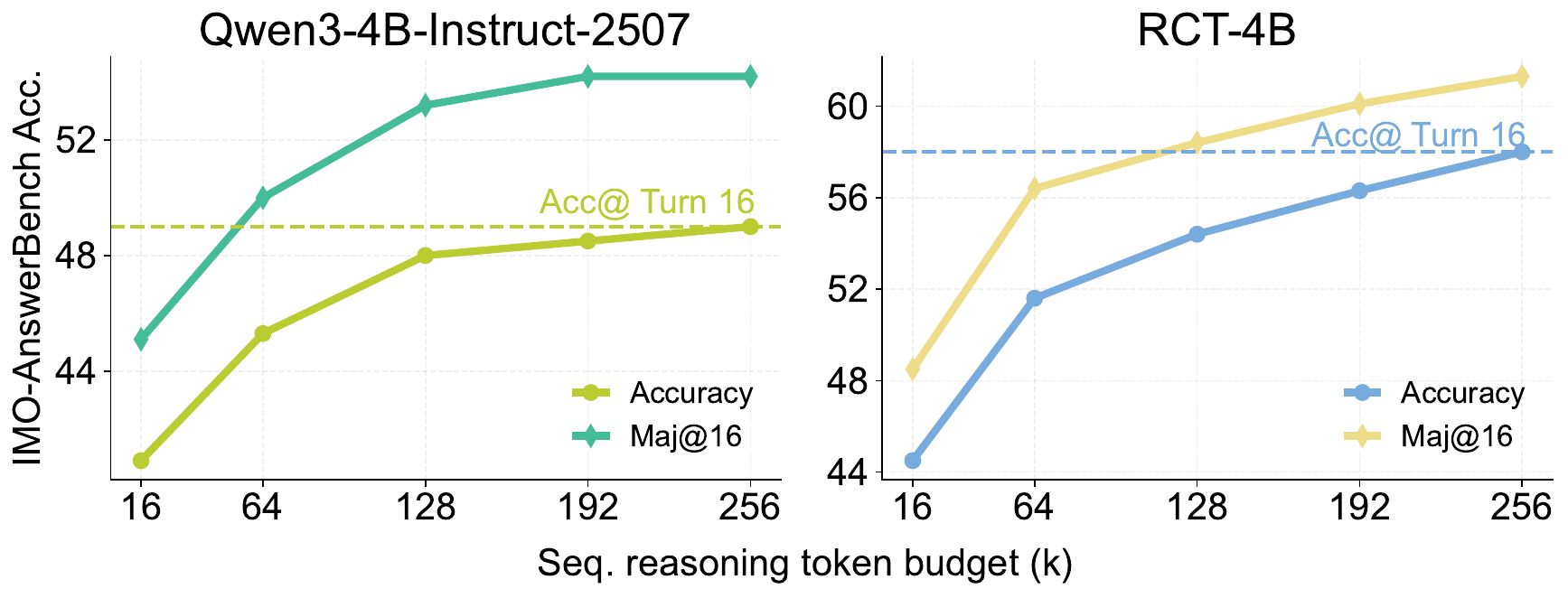}
    \vspace{-0.2cm}
    \caption{\footnotesize{\textbf{Accuracy and Maj@16 against sequential reasoning token budget for Qwen3-4B-Instruct-2507 and \texttt{RCT-4B} on IMO-AnswerBench}. While majority voting can be used to improve \methodname{}, we find that utilizing compute to increase ``depth'' through \methodname{} is more effective than increasing ``breadth'' by taking majority vote over more parallel samples.}}
    \label{fig:maj_vote}
    \vspace{-0.1cm}
\end{figure}

In Figure~\ref{fig:maj_vote}, we plot accuracy and Maj@16 against sequential reasoning token budgets using \methodname{} decoding (true reasoning token budget for Maj@16 is the sequential reasoning token budget multipled by 16). While majority voting can be used to improve \methodname{}, we find that utilizing compute to increase ``depth'' through \methodname{} is more effective than increasing ``breadth'' by taking majority vote over more parallel samples for our tested value of $k$: in other words, Maj@16 performance at 16k sequential reasoning token budget (giving a true reasoning token budget of 256k) is significantly worse than accuracy at 256k tokens with \methodname{}. This applies both for the base Qwen3 model and for \texttt{RCT-4B}.

\section{Motivating our Choices of $H_R$}\label{app:H_R_choice}

In this section, we motivate the choices of $H_R$ (autoregressive decoding maximum token budget) we use throughout this work. For \methodname{} decoding, $H_R$ determines the length of individual reasoning traces within each turn. As discussed in Section~\ref{subsec:rc_eval}, we generally choose $H_R$ to be $H_\text{train}$. Unfortunately, the exact value of $H_\text{train}$ is typically not made public, so we must estimate it through the termination length statistics of the model: if the model generally terminates its reasoning within some length $L$, then we can reasonably say that $L \approx H_\text{train}$. 

Regardless, the general idea is that increasing the reasoning token budget beyond $L$ will not yield any gains in performance with autoregressive decoding, as the model will simply not generate anything longer than this. For our base Qwen3-4B-Instruct-2507 model, we set $H_R = 16\text{k}$, which yields a very high termination rate of 99.17\% (see Table~\ref{tab:termination_rates}. Our trained \texttt{RCT-4B} model also attains a very high termination rate of 98.75\%, indicating that our training has not resulted in increased repetitiveness or undue verbosity. For our autoregressive decoding baselines, we choose the maximum token budget based on the values recommended on each model's Hugging Face model card. As we see from Table~\ref{tab:termination_rates}, reasoning traces generated at these lengths do indeed overwhelmingly terminate successfully.

A note on input context windows:  many models have a stated maximum context windows that are very large. For example, Qwen3-4B-Instruct-2507 has a stated maximum context window of 262,144 tokens. However, we note that they almost never generate reasoning traces of lengths greater than 16k tokens: see Table~\ref{tab:termination_rates}. This is likely because the models were post-trained to generate outputs of up to 16k tokens in length: the 262,144 context window is only utilized for processing long-context \emph{inputs}.

\begin{figure}[t]
    \centering
    \includegraphics[width=0.6\textwidth]{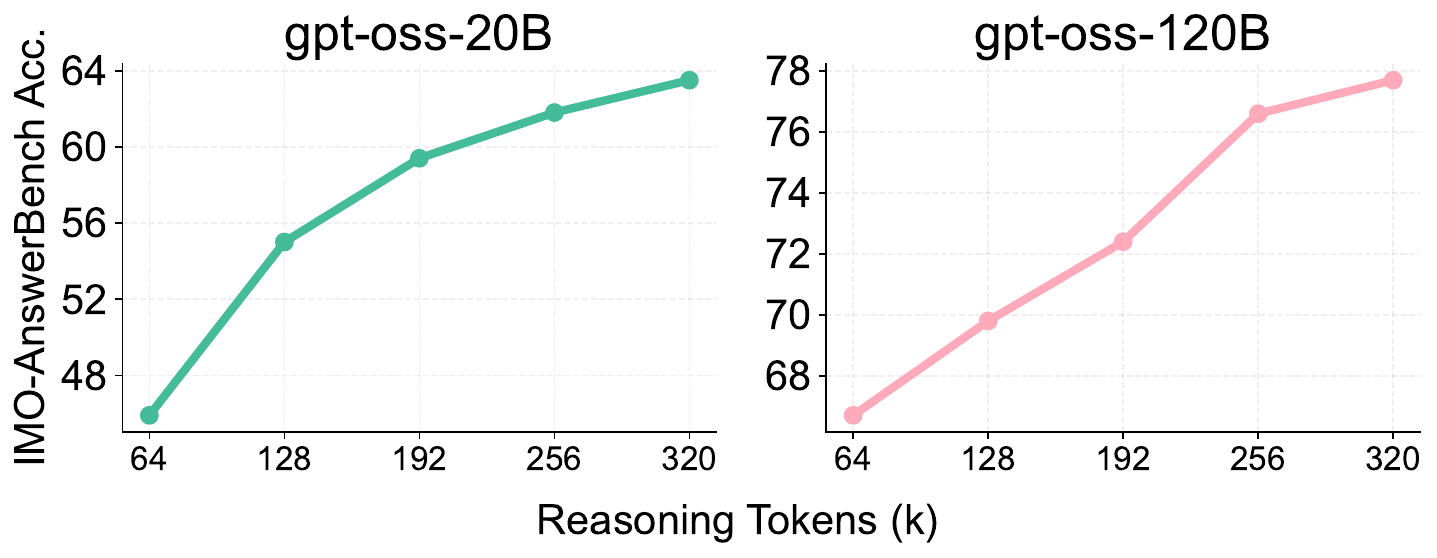}
    \vspace{-0.3cm}
    \caption{\footnotesize{\textbf{Performance of gpt-oss models with \methodname{} decoding (no training) on IMO-AnswerBench.} We use $H_R = 64\text{k}$ to adjust for the models' larger output lengths, and set reasoning effort to ``high'' for generation and ``medium'' for summarization. Both models benefit from extrapolation through \methodname{}. We evaluate answer correctness using Gemini-3-Flash as an LLM judge.}}
    \label{fig:gptoss_rc}
    \vspace{-0.2cm}
\end{figure}

\begin{table}[h]
\centering
\footnotesize
\begin{tabular}{l| cc}
\hline
\textbf{Model} & $H_R$ & \textbf{Termination Rate (\%)} \\
\hline
Qwen3-4B-Instruct-2507 & 16k & 99.17 \\
Qwen3-4B-Thinking-2507 & 81k & 100.00 \\
Polaris-4B & 90k & 99.79 \\
Qwen3-4B-Instruct-2507 + Std. RL @32k & 32k & 99.17 \\
RCT-4B & 16k & 98.75 \\
\hline
\end{tabular}
\caption{\footnotesize{\textbf{Termination rates for different models on HMMT 2025.} We determine termination by the presence of the \texttt{boxed\{\}} pattern in the model output.}}
\label{tab:termination_rates}
\end{table}

\section{\methodname{} with gpt-oss}\label{app:gptoss}

Please see results showing the efficacy of our approach on this model family in Figure~\ref{fig:gptoss_rc}.

\section{Limitations of \methodname{} and Directions for Future Work}\label{app:limitations}

While \methodname{} training yields strong empirical results, our method has several limitations that we outline in this section. We hope this discussion provides useful directions for future work.

\begin{wrapfigure}{r}{0.35\textwidth}
\vspace{-0.5cm}
  \begin{center}
    \includegraphics[trim=0 0 0 0, width=0.99\linewidth]{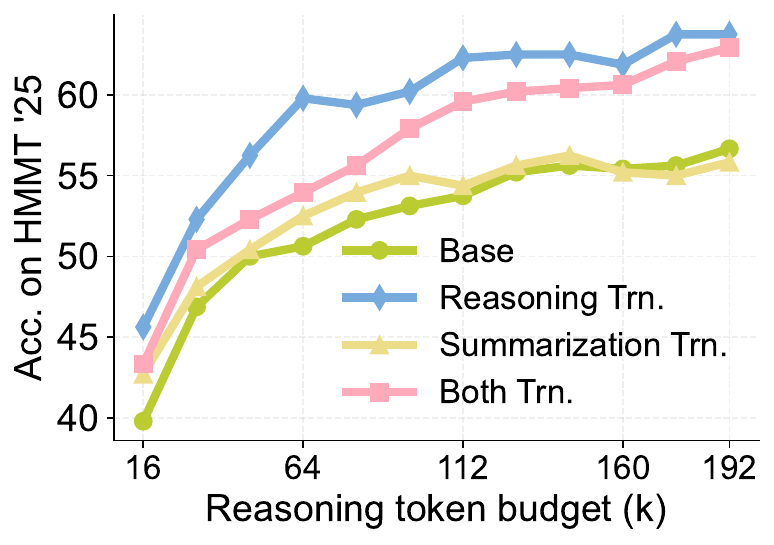}
  \end{center}
  \vspace{-0.5cm}
  \caption{\footnotesize{\textbf{Training for summarization generation using objective~\ref{eq:summary_gen_obj} hurts performance.} Training this way either in isolation or in combination with the usual summary-conditioned reasoning objective negatively impacts performance.}}
    \label{fig:summ_training}
    \vspace{-0.9cm}
\end{wrapfigure}
\textbf{\methodname{} training does not optimize summary generation.} Our training focuses exclusively on summary-conditioned reasoning, based on the observation that this is the primary performance bottleneck while summary generation is inherently easier for pre-trained and instruction-tuned models. We validated this assumption through preliminary experiments where we assigned rewards to summary generation based on the proportion of subsequent reasoning traces (conditioned on the summary) that produced correct answers. More formally, we optimized the following objective:
\begin{align}
    \max_{\pi_\theta}~ & \mathbb{E}_{\substack{\x, \y \sim \mathcal{D}_{\text{train}} \\ t \sim U[1, \mathrm{T}_{\mathrm{train}}]}} \left[ \mathbb{E}_{\z_S^{(t)} \sim \pi_\theta(\cdot | \mathcal{I}_S, \x, \z_R^{(t)}, \z_S^{(t-1)})} \left[ \frac{1}{K}\sum_{k=1}^K r(\y, \z^{(t+1)}_{R, k}) \right] \right], \nonumber\\
    &\text{where}\quad \z^{(t+1)}_{R, k} \sim \pi_{\theta}(\cdot|\mathbf{x}, \z_S^{(t)}) ~~~\text{ for } k=1,\ldots,K.
    \label{eq:summary_gen_obj}
\end{align}
We tested this both in isolation and in combination with our usual summary-conditioned generation objective, using the same hyperparameters as in Section~\ref{sec:rc_training} for Stage I training. As illustrated in Figure~\ref{fig:summ_training}, optimizing for summary generation only (``Summarization Trn.'') hurts the efficacy of training, such that the resulting model is no better than the base model. Optimizing for both summary generation and summary-conditioned generation (``Both Trn.'') improves performance relative to the base model but hurts performance relative to the model trained only for summary-conditioned generation (``Reasoning Trn.''). These findings motivate the design of our main approach.

We attribute these results to difficulties in credit assignment for summary generation. Even when the model generates faithful, informative summaries, it receives zero reward if subsequent reasoning fails to solve the problem, which may occur simply because the problem is too difficult to solve in a single turn and not because the summary is poor. This misalignment between summary quality and reward signal makes it difficult to effectively train summarization, although we posit that doing so effectively could further improve \methodname{} performance. Addressing this likely requires alternative reward assignment schemes for summary generation, which we leave to future work.

\textbf{\methodname{} training uses myopic rewards.}
\methodname{} is trained by optimizing Equation~\ref{eq:train_obj} using outcome-based rewards assigned independently at each iteration. That is, the reasoning trace generated at turn $t$ receives reward based solely on its own correctness, without explicit credit assignment to future turns $t+1, t+2, \ldots$. As discussed in Section~\ref{sec:rc_training}, this design relies on the observation that training each step to produce a correct answer implicitly encourages the model to generate reasoning and summaries that contain information useful for subsequent iterations. When each summarization and summary-conditioned generation step makes monotonic progress toward a correct solution, such myopic rewards are well aligned with trajectory-level success. In this regime, summaries serve as sufficient representations of past reasoning, and improving per-step correctness also improves the quality of information available for future steps. Indeed, under the assumption that the model can reliably extract maximal useful information from a summary when needed, optimizing per-step correctness is sufficient to optimize the long-horizon objective over multiple iterations. Moreover, by conditioning training rollouts on these summaries, the model is explicitly trained to exploit accumulated information, further increasing the likelihood of success at subsequent steps. As a result, we can optimize correctness over the full reasoning trajectory despite only explicitly optimizing individual steps.

The main limitation of our approach is that the model is not incentivized to generate reasoning that is suboptimal for the current step but valuable later. For example, the model may benefit from exploring alternative solution strategies or collecting auxiliary information that only becomes useful later in the trajectory. Learning such far-sighted reasoning behaviors may be particularly important for very difficult problems requiring extensive in-context exploration. However, designing reward schemes that effectively encourage such multi-step contributions remains challenging, and we leave this direction to future work.

\textbf{Summarization-generation asymmetry is not present in all LLMs.} Our analysis in Section~\ref{subsec:rc_eval} reveals that \methodname{} yields the most benefits when the underlying model possesses strong summarization-generation asymmetry, and that instruction-following models generally possess this asymmetry whereas highly specialized reasoning models do not. This limits the kinds of models we can apply \methodname{} to. We propose several potential solutions to this problem. The first involves warmstarting the reasoning model for summarization and summary-conditioned generation, perhaps through distillation or SFT. This approach, however, may potentially alter the reasoning behavior of the model in a detrimental way. The second solution involves using a separate model to perform summarization generation, which we previously identified as a particularly difficult task for specialized reasoning models. This approach, however, would then require us to maintain two separate models, which could pose certain practical challenges.

\textbf{\methodname{} does not improve performance on all classes of reasoning problems.} While our experiments show that \methodname{} decoding and training improves model performance across mathematical and scientific reasoning benchmarks, we posit that not all classes of problem classes benefit from \methodname{}. One class of such problems are search-heavy problems, where the model must iterate through a large number of possible outcomes and select the optimal choice. The main issue here is that the \emph{redundancy} property no longer applies as strongly as before, as many tokens generated may be important as they document the search process and keep track of what has been tried. Summarizing such traces risks discarding important information that may reduce test-time performance on the search task.

\emph{\textbf{When is \methodname{} helpful?}} To understand the classes of problems for which \methodname{} is most effective, it is useful to conceptualize the solution space as a graph, where nodes represent conclusions, intermediate results, or other salient states, and edges represent logical transitions between them. \methodname{} is particularly well-suited to problems whose solution graphs are \emph{clique-like}: nodes tend to form loosely connected clusters that can be summarized compactly, with relatively sparse connections between clusters. Mathematical and scientific reasoning often exhibit this structure, as progress can be decomposed into distinct conceptual advances that admit concise summaries. 

In such settings, \methodname{} enables effective reasoning by allowing each iteration to explore different regions of the graph, while maintaining long-term progress by summarizing and tracking information within individual clusters. In contrast, for search-heavy problems this clique structure is largely absent, and effective reasoning requires tracking a large number of individual nodes encountered during exploration. In these cases, summaries must enumerate prior states rather than abstract them, causing limited-length summaries to quickly become overwhelmed and reducing the effectiveness of iterative summarization.

In addition to scientific and mathematical reasoning, \methodname{} may also be helpful on tasks where actions yield environment feedback that is noisy and can benefit from summarization (e.g. coding with interpreter feedback). In this case, summaries may be used to keep track of environment state, as has been explored in related work~\citep{zhou2025mem1learningsynergizememory}. Unlike these works, we focus primarily on creating abstractions of reasoning rather than environment states, but these ideas are closely related and can likely be combined.

\section{Discussions on Computational Efficiency}\label{app:efficiency}

\subsection{Inference}
We begin by analyzing the computational efficiency of \methodname{} decoding compared to standard long-context autoregressive generation. We examine how \methodname{} scales with reasoning length and whether extrapolating via \methodname{} is more efficient than training models to natively handle larger token budgets through autoregressive decoding.

\begin{wrapfigure}{r}{0.35\textwidth}
\centering
\vspace{-0.3cm}
\includegraphics[width=0.99\linewidth]{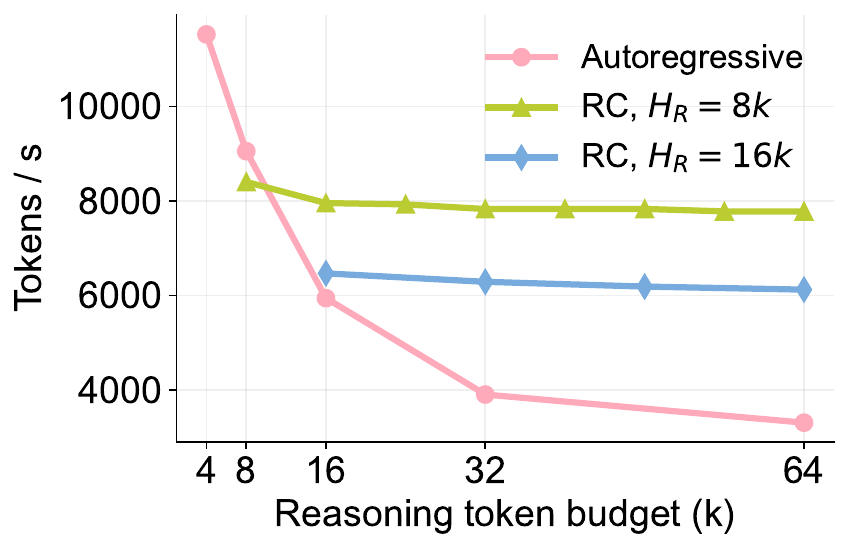}
\vspace{-0.6cm}
\caption{\footnotesize{\textbf{Plot of decoding throughput against reasoning token budget.} \methodname{} decoding throughput remains constant as reasoning token budget, whereas throughput for standard autoregressive decoding decreases.}}
\label{fig:throughput_plot}
\end{wrapfigure}
\textbf{Notation and definitions}. Let $C$ be the input problem length, and let $N$ be the maximum generation length for standard autoregressive decoding. Under \methodname{} decoding, the model proceeds for $\mathrm{T}$ turns, generating at each turn a reasoning statement of length $\le  H_R$ followed by a summary of length $\le H_S$, with $H_R \gg H_S$. Our analysis focuses on decoder-only transformers with KV-cached decoding, where for long contexts, attention computation dominates and scales linearly with current context length.

\textbf{Standard Long-Context Generation}. In standard autoregressive decoding, the model generates $N$ tokens in a single trajectory, with the context growing from length $C$ to $C+N$. With KV caching, the incremental cost of generating the $i$-th token scales linearly with the context length $C+i$. Summing over all tokens, the total attention-dominated inference cost (IC) scales as
\begin{align}
\text { IC }_{\text {standard}} \propto \sum_{i=1}^N(C+i)=N C+\frac{1}{2} N(N+1)=\Theta(N(C+N))
\end{align}

\textbf{\methodname{} Inference}. In \methodname{}, each reasoning step is conditioned only on the original prompt and the current summary, rather than the full previous chain-of-thought. As a result, the effective context length within each turn is bounded by approximately $C+H_S+H_R$, and \textbf{does not grow across turns}. Across $\mathrm{T}$ turns, the total inference compute is therefore
\begin{align}
\text { IC }_{\methodname{}} \propto T \cdot H_R\left(C+H_S+H_R\right)
\end{align}

\textbf{Inference Speedup}. For a fixed effective reasoning budget $N=\mathrm{T} H_R$, the inference speedup of \methodname{} is approximately
\begin{align}
\text { Speedup }= \frac{\text {IC}_\text{standard}}{\text { IC }_{\methodname{}}}\approx \frac{C+T H_R}{C+H_S+H_R}\approx T.
\end{align}
as $\mathrm{T} H_R \gg C$ and $H_R \gg H_S$. Therefore, to reach 
$N=\mathrm{T} H_R$ effective reasoning tokens, \methodname{} can be $\approx \mathrm{T}$ times cheaper than autoregressive decoding in attention-dominated regimes.

\textbf{Empirical study.} 
We conduct experiments to study the computational efficiency of \methodname{}. We run \methodname{} decoding using Qwen3-4B-Instruct-2507 and standard autoregressive decoding using Qwen3-4B-Thinking-2507 at different reasoning token budgets, logging throughput on HMMT 2025 (30 prompts) with 8 parallel rollouts. For the \methodname{} runs, we experiment with $H_R \in \{8\text{k}, 16\text{k}\}$. We plot decoding throughput against reasoning token budget in Figure~\ref{fig:throughput_plot}. This plot demonstrates that autoregressive decoding throughput rapidly decreases as token budgets increase, whereas \methodname{} decoding throughput remains constant. This is expected because \methodname{} maintains bounded context length across turns even as the effective reasoning horizon grows. \methodname{} therefore proves substantially more efficient than autoregressive decoding despite our use of a highly optimized inference engine for the latter and a naive implementation for the former: see Appendix~\ref{app:hardware_hparams} for implementation and hardware details.

\subsection{Training}

\textbf{Standard long-context RL baseline}. In both standard RL training, we perform on-policy RL (e.g., GRPO) with batch size (problems per step) = $B$ and samples per problem = $K$ (GRPO group size). Then each step generates roughly $B \cdot K \cdot N$ tokens, where $N$ is the sequence length. Since the attention cost scales with the growing context length, the forward generation compute scales as:
\begin{align}
\text { GenCompute }_{\text {standard }} \propto B \cdot K \cdot N(C+N) .
\end{align}
Including backward and optimizer computation introduces a constant multiplicative factor $\gamma$, yielding
\begin{align}
    \text { TrainCompute }_{\text {long }} \approx \gamma \cdot B \cdot K \cdot N(C+N) .
\end{align}
When $N$ is large, this scales quadratically with the rollout horizon.

\textbf{\methodname{} Training}. \methodname{} training separates trajectory construction from policy optimization. Each training step consists of: (1) summary-trajectory generation: The model runs \methodname{} for $\mathrm{T}_{\text {train }}$ turns to produce a sequence of summaries; (2) policy optimization: from this trajectory, $N_\text{summ}$ summaries are sampled, and for each summary, $K$ reasoning rollouts of length at most $H_R$ are generated and optimized via GRPO.

The total forward generation compute per training step scales as
\begin{align}
\text { GenCompute }_{\mathrm{\methodname{}}} \propto B \cdot\left(\mathrm{T}_{\text {train }}+K N_\text{summ}\right) \cdot H_R\left(C+H_S+H_R\right) .
\end{align}
Including backward and optimizer cost yields
\begin{align}
\text { TrainCompute }_{\mathrm{\methodname{}}} \approx \gamma \cdot B \cdot\left(\mathrm{T}_{\text {train }}+K N_\text{summ}\right) \cdot H_R\left(C+H_S+H_R\right) .
\end{align}
Crucially, all optimized rollouts remain bounded by length $H_R$, regardless of the total effective reasoning horizon supported at inference time.

\textbf{Training-Time Scaling Comparison}. To reach an effective horizon $N=\mathrm{T}_{\text {target }} H_R$, standard long-context RL training incurs compute scaling annroximatelv as
\begin{align}
\text { TrainCompute }_{\text {standard }} \propto B \cdot K \cdot \mathrm{T}_{\text {target }}^2 H_R^2
\end{align}
while \methodname{} training scales as
\begin{align}
\text { TrainCompute }_{\mathrm{\methodname{}}} \propto B \cdot\left(\mathrm{T}_{\text {train }}+K N_\text{summ}\right) \cdot H_R^2
\end{align}
Thus, the relative cost satisfies
\begin{align}
\frac{\text { TrainCompute }_{\mathrm{\methodname{}}}}{\text { TrainCompute }_{\mathrm{standard}}} \approx \frac{\mathrm{T}_{\mathrm{train}}K N_\text{summ}}{K \cdot \mathrm{T}_{\mathrm{target}}^2} .
\end{align}
This highlights a key advantage of \methodname{}: naively increasing rollout length leads to quadratic growth in training cost, whereas \methodname{} decouples the optimized rollout length from the effective reasoning horizon. By using summaries and replay, \methodname{} enables training policies that generalize to very long reasoning horizons without incurring prohibitive quadratic costs during optimization.

\subsection{Inference KV-Cache Memory}
The KV cache memory footprint for autoregressive decoding scales linearly with the context length:
\begin{align}
    \text { Memory }_{\text{standard}} \propto C+N,
\end{align}
while for \methodname{}, it is bounded by the maximum within-turn context length:
\begin{align}
\text { Memory }_{\methodname{}} \propto C+H_S+H_R,
\end{align}
which is independent of $\mathrm{T}$. Putting these together, \methodname{} requires $\sim T \times$ lower KV memory at the same effective reasoning horizon:
$$
\frac{\text{Memory}_{\text {standard}}}{\text{Memory}_{\methodname{}}} \approx \frac{C+T H_R}{C+H_S+H_R} \approx T .
$$

\section{Details for Test-Time Scaffold Experiments}\label{app:scaffold_details}

\subsection{Recursive Self-Aggregation}
In Section~\ref{sec:scaffolds}, we also experiment with incorporating \methodname{} into RSA~\citep{venkatraman2025recursiveselfaggregationunlocksdeep}, a scaffold that iteratively refines solutions through sampling and aggregation. In its original form, the algorithm begins by sampling $M$ solutions from scratch (conditioned only on the problem). Then, in each subsequent iteration, the algorithm creates $M$ new solutions by randomly sampling $k$ candidates from the current pool of solutions (with replacement) and prompting the model to aggregate them into a single improved solution. Over the $\mathrm{T}_\text{RSA}$ successive loops, solutions compound recursively: aggregated outputs become inputs for the next round, progressively eliminating errors and reinforcing correct solutions while maintaining a constant population of $M$ solutions.

We incorporate \methodname{} into RSA by replacing \textbf{(1)} the initial solution generation step and \textbf{(2)} subsequent refinement steps with  \methodname{} decoding. We begin the refinement step by treating the aggregated solution as a summary that we condition on for the first step of \methodname{} refinement. For our experiments in Section~\ref{sec:scaffolds}, we use $k = 2$, $M = 8$, and $\mathrm{T}_\text{RSA} = 10$, and for the experiment incorporating \methodname{} decoding, we set the number of \methodname{} steps as $\mathrm{T} = 8$.

\subsection{DeepseekMath Agent}
We also experiment with a test-time scaffold we call DeepseekMath Agent (DSM Agent). This is adapted from the scaffold used in \citet{shao2025deepseekmathv2selfverifiablemathematicalreasoning} to improve the ability of LLMs to generate proofs for mathematical reasoning problems. 

At a high level, the DSM Agent implements a Generate-Verify-Refine loop that uses self-verification to iteratively improve solutions. It begins by generating an initial pool (of size $n_g$) of candidate solutions, and then verifies each solution using $n_v$ self-verification attempts per solution (assigning scores of 0.0 for major errors, 0.5 for minor issues, 1.0 for correct), with the final verification score determined by averaging over the $n_v$ scores. In each of the subsequent refinement iterations, the algorithm selects the highest-scoring solutions and refines them using feedback from their lowest-scoring verifications. Refined solutions are added to the growing pool and re-verified, with this process repeating until either \textbf{(1)} a perfect score is achieved, or \textbf{(2)} the maximum $\mathrm{T}_\text{DSM}$ iterations are reached. At the end, the algorithm returns the highest-scoring solution as the final answer.

We incorporate \methodname{} into DSM Agent by replacing the initial solution generation step with \methodname{} decoding, with the aim of improving the quality of the initial pool of candidates. For our experiments in Section~\ref{sec:scaffolds}, we use $n_g = 8$, $n_v = 4$, and $\mathrm{T}_\text{DSM} = 6$, and for the experiment incorporating \methodname{} decoding, we set the number of \methodname{} steps as $\mathrm{T} = 8$.

\section{Overview of GRPO}\label{app:grpo_overview}

GRPO optimizes the following objective:
\begin{align}
\label{eq:grpo}
    \mathcal{J}(\theta) = \mathbb{E}_{\x, \y \sim \mathcal{D}_{\text{train}}} 
    \mathbb{E}_{\z_i \sim \pi_{\theta}(\cdot \mid \x)}
    \left[
    \frac{1}{K} \sum_{i=1}^{K} \min \left[ \frac{\pi_\theta(\z_i \mid \x)}{\pi_{\text{old}}(\z_i \mid \x)} A_i, \text{clip}\left(\frac{\pi_\theta(\z_i \mid \x)}{\pi_{\text{old}}(\z_i \mid \x)}, 1-\epsilon, 1+\epsilon\right) A_i \right]
    \right].
\end{align}
Here, $\z_i$ denotes the $i$th of $K$ independently sampled rollouts (which taken together form a ``group''), and $A_i$ denotes the GRPO advantage, which is computed directly from the rewards as $A_i = \frac{r_i - \mathrm{mean}(\mathbf{r})}{\mathrm{std}(\mathbf{r})}$, with the mean and standard deviation calculated over group rewards.

Some intuitions behind GRPO:
\vspace{-0.8em}
\begin{itemize}
    \item For a fixed input $\x$, GRPO assigns advantages to each rollout $\z_i$ relative to the other $K$ samples in the group, so updates depend on whether $\z_i$ is better or worse than its peers rather than on absolute reward values. In the case of \methodname{} training, the $K$ parallel rollouts are sampled under the same prompt and summary combination, so we assign higher advantages to summary-conditioned reasoning traces that are better able to leverage the summary to attain the correct answer.
    \item Normalizing advantages by the group mean and standard deviation stabilizes gradients and makes updates invariant to the overall reward scale across different inputs.
    \item The clipped ratio $\pi_\theta(\z_i \mid \x) / \pi_\text{old}(\z_i \mid \x)$ retains PPO’s~\citep{schulman2017ppo} trust-region approach, preventing any single high-advantage $\z_i$ from applying overly large updates.
\end{itemize}

\section{Dataset Construction Details}\label{app:data_construction}

We construct our training datasets by following some of the principles outlined in \citet{Polaris2025}. Specifically, we sample problems in a way that ensures our dataset maintains reasonable difficulty given our model. We begin by sampling problems from the AceReason-Math~\citep{chen2025acereason} dataset ($\sim 50\text{k}$ problems) and solving them with Qwen3-4B-Instruct-2507, setting $K = 64$. We then evaluate these solutions and assign a reward score to each problem based on the average number of correct solutions our model generates. These scores are used for weighted sampling: we discard all samples that attain a score of 0.7 or greater, and downsample problems with other reward scores to attain the ``J-shaped'' reward curve described in Figure 2 of~\citet{Polaris2025}. This procedure yields a dataset of around 5.7k samples, which we take as our Stage I training set.

After Stage I training, our model improves and so we rebalance our training dataset such that it maintains the ``J-shaped'' reward curve. We reannotate our Stage I dataset with the Stage I model (with standard autoregressive decoding) and once again remove samples that attain reward scores of 0.7 of greater. We then inject $\sim 500$ difficult (zero-reward) problems from the DAPO~\citep{yu2025dapoopensourcellmreinforcement} dataset (as determined via annotation with the base model) and ensure that our Stage II dataset contains challenging problems.

\section{Iterative Decoding Baseline Details}\label{app:iter_baselines}
In this section, we describe in detail the self-verification and self-refinement iterative decoding baselines that we compare \methodname{} against. The purpose of these baselines is to help us separate out the impact of our summarize-generate routine from the impact of using iterative decoding. As such, these baseline methods do not utilize the summarization-generation asymmetry, and instead act directly on the reasoning trace generated by the model, as is common in iterative decoding methods and test-time scaffolds~\citep{shao2025deepseekmathv2selfverifiablemathematicalreasoning, kumar2024traininglanguagemodelsselfcorrect, qu2024recursiveintrospectionteachinglanguage}.

\subsection{Inference} 

Concretely, let $\x$ denote the input prompt and let $t \in \mathbb{N}$ index the decoding turn. Unlike \methodname{}, these baseline methods maintain only a reasoning trace $\z_R^{(\mathrm{t})}$ at each turn, with no separate summarization step. At each turn, the reasoning trace $\z_R^{(\mathrm{t})}$ is generated under a fixed token budget $H_R$ (we use the same $H_R = 16\text{k}$ as in our \methodname{} experiments).

For \textbf{self-refinement}, decoding proceeds by alternately generating reasoning traces and prompting the model to refine them. At each turn $\mathrm{t}$, we sample:
\begin{align}
\z_R^{(\mathrm{t})} &\sim \pi_\theta\!\left(\cdot \mid \mathcal{I}_{\text{refine}}, \x, \z_R^{(\mathrm{t}-1)}\right),
\end{align}
where $\mathcal{I}_{\text{refine}}$ instructs the model to improve upon its previous reasoning trace, and $\z_R^{(0)}$ is initialized as the empty string.

For \textbf{self-verification}, the model is prompted to first verify its previous attempt before generating a correction. At each turn $\mathrm{t}$, we sample:
\begin{align}
\z_R^{(\mathrm{t})} &\sim \pi_\theta\!\left(\cdot \mid \mathcal{I}_{\text{verify}}, \x, \z_R^{(\mathrm{t}-1)}\right),
\end{align}
where $\mathcal{I}_{\text{verify}}$ instructs the model to verify whether its previous reasoning is correct and, if not, to provide a corrected solution. See Figures~\ref{fig:refine_prompt} and \ref{fig:verify_prompt} for $\mathcal{I}_{\text{refine}}$ and $\mathcal{I}_{\text{verify}}$.

After $\mathrm{T}$ decoding turns, the final output is given by $\z := \z_R^{(\mathrm{T})}$ for both methods. The key distinction from \methodname{} is that these baselines condition on the full previous reasoning trace $\z_R^{(\mathrm{t}-1)}$ rather than a compressed summary. As such, the model must conditionally generate from sequences up to $2H_R$ in length.

\subsection{Training}
Training follows a similar scheme to \methodname{} training, except that we generate rollouts using our baseline iterative decoding methods instead of \methodname{} decoding. The idea here is to assess whether utilizing the summarization-generation gap enables us to achieve better performance through training, or whether our iterative training strategy on its own is sufficient to attain significant gains.

More formally, at any given point in training, we run the iterative decoding algorithm for $\mathrm{T}_\text{train}$ turns for each problem $\x$ in a training batch. We collect the reasoning traces generated from these rollouts $\z_R := (\z_R^{(1)}, \dots, \z_R^{(\mathrm{T}_\mathrm{train})})$ and then uniformly sample $N_{\text{trace}} \leq \mathrm{T}_{\mathrm{train}}$ unique traces per problem. We then generate $K$ reasoning traces conditioned on each sampled trace. We assign rewards based on correctness and compute advantages over these $K$ samples. Formally, the objective can be written as:
\begin{align}
    \max_{\pi_\theta}~~~ & \mathbb{E}_{\x, \y \sim \mathcal{D}_{\text{train}}, t \sim U[1, \mathrm{T}_{\mathrm{train}}]} \left[ \mathbb{E}_{\z^\prime \sim \pi_{\theta}(\cdot|\mathbf{x}, \z_R^{(t)})} \left[r(\y, \z^\prime)\right]\right], \quad |\z^\prime| \leq H_R \nonumber\\ 
    &\text{where}\quad(\z_R^{(1)}, \dots, \z_R^{(\mathrm{T}_\mathrm{train})})
    \sim \text{IterativeDecoding}(\pi_\theta; \x).
    \label{eq:baseline_train_obj}
\end{align}

We adopt the same training hyperparameters as for \methodname{} training in Section~\ref{sec:rc_training}.

\section{Additional Iterative Decoding Comparisons}\label{app:additional_baselines}

We evaluate two additional prompting-only iterative decoding baselines beyond the self-refinement and self-verification methods described in Appendix~\ref{app:iter_baselines}. Rather than conditioning directly on the full reasoning trace, these baselines modify the trace before conditioning on them for subsequent generation.

\subsection{Budget Forcing}

\begin{table}[h]
\centering
\footnotesize
\begin{tabular}{l | c c c}
\toprule
& \textbf{AIME 2025} & \textbf{HMMT 2025 (Nov)} & \textbf{FrontierScience} \\
\midrule
Qwen3-4B-Instruct-2507 [16k] & 46.0 & 39.8 & 23.3 \\
Qwen3-4B-Instruct-2507 + \methodname{} & 59.4 & 56.7 & 29.5 \\
\textbf{\texttt{RCT-4B} + \methodname{}} \textcolor{red}{\textbf{\textit{(Ours)}}} & 74.9 & \textbf{66.3} & \textbf{34.1} \\
Budget Forcing~\citep{muennighoff2025s1simpletesttimescaling} & 56.7 & 46.7 & 20.1 \\
\bottomrule
\end{tabular}
\caption{\footnotesize{\textbf{Comparison of budget forcing with Qwen3-4B-Instruct-2507 against the autoregressive baseline and against \methodname{}.} Budget forcing yields modest gains on mathematical reasoning benchmarks, but degrades performance on FrontierScience vs. the autoregressive baseline. \methodname{}-based methods outperform budget forcing on all benchmarks.}}
\label{tab:s1}
\end{table}

The first baseline we evaluate is a budget forcing approach, inspired by \citet{muennighoff2025s1simpletesttimescaling}. In this approach, we repeatedly append \texttt{``Wait, let me continue thinking''} after termination to elicit additional reasoning. We use $H_R = 16\text{k}$, and report our results in Table~\ref{tab:s1}.

Table~\ref{tab:s1} shows that budget forcing yields modest gains on mathematical reasoning benchmarks relative to autoregressive decoding but degrades FrontierScience performance. Both \methodname{}-based methods substantially outperform budget forcing across all benchmarks. Analyzing output traces reveals two failure modes. First, budget forcing rarely changes the final answer after the first few turns, with continuations quickly reducing. We hypothesize that simply appending \texttt{``Wait...''} is insufficient encouragement for the model to engage in meaningful in-context exploration, unlike conditioning on entire summaries of past reasoning. Second, excessive forcing steps often produce degenerate behavior characterized by substantial repetition, especially in later turns. We attribute these issues to repeated \texttt{``Wait...''} prompts creating highly unnatural, out-of-distribution prefixes, especially after the model has already returned a final answer. In contrast, conditioning on summaries maintains in-distribution prefixes, yielding better performance.

\subsection{Delethink}

\begin{wrapfigure}{r}{0.4\textwidth}
\vspace{-0.9cm}
  \begin{center}
    \includegraphics[trim=0 0 0 0, width=0.99\linewidth]{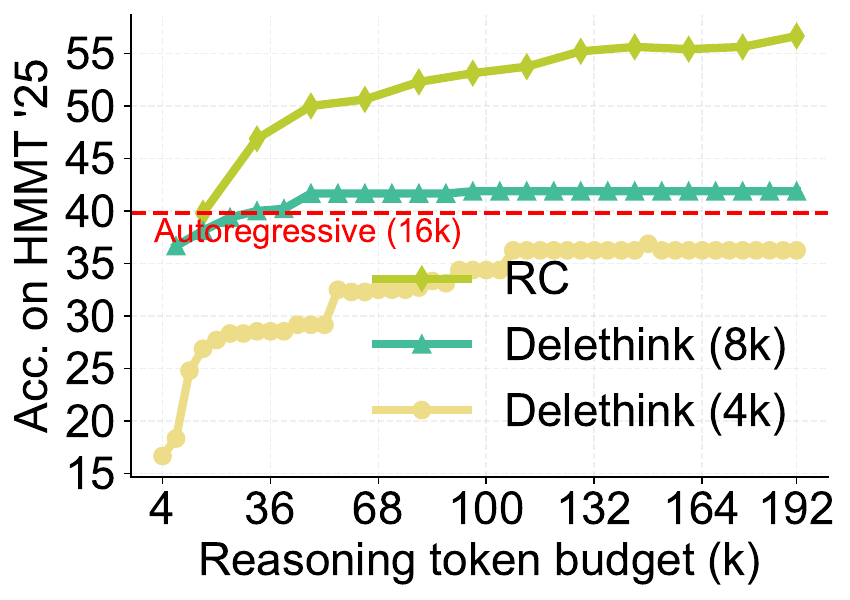}
  \end{center}
  \vspace{-0.7cm}
  \caption{\footnotesize{\textbf{Performance as a function of reasoning token budget, comparing \methodname{} with Delethink~\citep{aghajohari2025markovianthinkerarchitectureagnosticlinear}.} We use Qwen3-4B-Instruct-2507 for both experiments, without any additional training. \methodname{} outperforms Delethink across all token budgets.}}
    \label{fig:delethink_baseline}
    \vspace{-0.3cm}
\end{wrapfigure}
Our second baseline is a prompting-only version of Delethink~\citep{aghajohari2025markovianthinkerarchitectureagnosticlinear}. At each iteration, Delethink generates $H_R$ tokens of reasoning conditioned on the final $H_\text{chunk}$ tokens from the previous iteration, continuing until EOS or reaching maximum iterations. This approach enables long-horizon RL training by chunking reasoning into segments, which circumvents the practical difficulties of standard long-context RL (a goal also shared by \methodname{} training: see Appendix~\ref{app:efficiency}).

We compare \methodname{} with the base Qwen3-4B-Instruct-2507 against Delethink using the same model. We evaluate Delethink with $H_R \in \{4\text{k}, 8\text{k}\}$, setting $H_\text{chunk}  = H_R/2$, and report our findings in Figure~\ref{fig:delethink_baseline}. We do not experiment with $H_R = 16\text{k}$ because the model overwhelmingly terminates its outputs at this value of $H_R$, preventing Delethink from yielding improvements at larger token budgets because the model simply will not continue generating. We find that Delethink generally improves performance as token budget increases, but plateaus near the performance of the autoregressive baseline (note that these results are consistent with findings in \citet{aghajohari2025markovianthinkerarchitectureagnosticlinear}). \methodname{}, on the other hand, improves significantly over the autoregressive decoding baseline, even without any \methodname{}-specific training.

We attribute this discrepancy to two factors. First, Delethink terminates after generating a final answer, which limits continued reasoning. In contrast, \methodname{} explicitly encourages verifying and refining prior answers, because subsequent reasoning is driven by the produced summary from the previous iteration. We find that this process drives significant performance gains. Second, Delethink conditions generation on only the prompt and the last $H_\text{chunk}$ reasoning tokens from the previous iteration, creating out-of-distribution prefixes for the base model. \methodname{} instead conditions generation on structured summaries of past reasoning, which, for models with strong instruction-following, is considerably more in-distribution. In particular, we hypothesize that on difficult problems, once a model has ceased to make meaningful progress within a chain-of-thought, conditioning on a short carryover from the previous iteration is often insufficient to induce reasoning that continues to make progress in the next turn. In contrast, \methodname{} partially alleviates this issue by conditioning on explicit summaries, which frequently prompt the model to restart, explore alternative strategies, or verify prior attempts. This difference may help explain why Delethink-style approaches tend to plateau, whereas \methodname{} continues to improve with additional iterations.

\section{Hardware, Hyperparameters, and Implementation Details}\label{app:hardware_hparams}

\begin{table}[h]
\centering
\caption{\footnotesize{\textbf{Training hyperparameters for all training experiments.}}}
\footnotesize
\begin{tabular}{lc}
\toprule
\textbf{Hyperparameter} & \textbf{Value} \\
\midrule
Learning rate & $1 \times 10^{-6}$ \\
KL loss coefficient & $0.001$ \\
Entropy loss coefficient & $0.0$ \\
Training batch size & $64$ \\
Minibatch size & $32$ \\
Clip range (low) & $0.2$ \\
Clip range (high) & $0.28$ \\
Gamma ($\gamma$) & $1.0$ \\
Lambda ($\lambda$) & $1.0$ \\
Warmup schedule & Constant \\
Weight decay & $0.01$ \\
Inference Temperature & $1.0$ \\
Inference top-$p$ & $1.0$ \\
Optimizer & AdamW \\
\bottomrule
\end{tabular}
\label{tab:hyperparameters}
\end{table}

\textbf{Hardware.} We conduct training on a single node of 8$\times$H100 GPUs, and conduct inference on either a single node of 8$\times$H100 GPUs or on a single node of 4$\times$GH200 GPUs. For our inference efficiency experiments in Appendix~\ref{app:efficiency}, we use 4$\times$GH200 GPUs.

\textbf{Implementation Details.} We use verl~\citep{sheng2024hybridflow} with FSDP~\citep{zhao2023pytorchfsdpexperiencesscaling} for training, and vLLM~\citep{kwon2023efficient} for inference. We use a naive implementation of \methodname{} decoding for both inference and training rollout generation. At each turn, we generate reasoning traces and summaries in batches sequentially, waiting for all sequences in a batch to complete before progressing to the next turn. This is suboptimal because sequences finish at different times (due to varying lengths and early termination), causing idle GPUs to wait for the slowest sequence in each batch. A more efficient implementation would perform asynchronous decoding across turns, allowing faster sequences to progress immediately without waiting for slower ones. Despite our use of a suboptimal implementation of \methodname{}, we still empirically observe significant efficiency benefits at longer reasoning lengths (Appendix~\ref{app:efficiency}).

\textbf{Inference Hyperparameters.} We perform inference with $t = 1.0$ and $p = 1.0$ for \texttt{RCT-4B}. For Qwen3-4B-Instruct-2507, we use $t = 0.7$ and $p = 0.8$. For Qwen3-4B-Thinking-2507, we use $t = 0.6$ and $p = 0.95$. 

\section{Full Example}\label{app:add_example}

\textbf{Example Problem}

\begin{lstlisting}
Let ABCD and CEFG be squares such that C lies on segment DG and E lies on segment BC. Let O be the circumcenter of triangle AEG. Given that A, D, and O are collinear and AB = 1, compute FG.
\end{lstlisting}

\textbf{Example Summary}

\begin{lstlisting}
In this solution, the goal is to compute the side length FG of square CEFG given that ABCD and CEFG are squares with C on segment DG, E on segment BC, A, D, and O (the circumcenter of triangle AEG) are collinear, and AB = 1. The solution begins by placing square ABCD on the coordinate plane with A at (0,0), B at (1,0), C at (1,1), and D at (0,1). Point E lies on segment BC, so it is parameterized as (1,e) for e \in [0,1]. The square CEFG is then constructed with side CE, and both clockwise and counterclockwise rotations are considered to determine the position of point G. In the counterclockwise case, G is found at (2-e, 2e-1); in the clockwise case, G is at (e,1). The condition that C lies on segment DG is then tested: for both orientations, the only value of e that satisfies this (with C strictly on the segment) is e=1, which results in E = C and a degenerate square. This degeneracy is unacceptable, so the solution attempts to resolve the issue by using the collinearity of A, D, and the circumcenter O of triangle AEG as a primary constraint instead of the segment condition. 

In the counterclockwise orientation, the circumcenter of triangle A(0,0), E(1,e), and G(2-e,2e-1) is computed using perpendicular bisectors, and the condition that this circumcenter lies on line AD (the line x=0) is imposed. Solving the resulting equation leads to a cubic in e: 3e^3 -7e^2 +3e +1 = 0, which has roots at e=1 and e = \frac{2 \pm \sqrt{7}}{3}, only one of which lies in [0,1], namely e=1. Similarly, in the clockwise case, the same condition leads to a different cubic that also only allows e=1 as a valid solution. Despite multiple attempts to find a non-degenerate configuration, all solutions reduce to e=1, which implies that side length FG = |CE| = 0, a degenerate case. Given that this configuration fails to yield a nontrivial solution and that all attempts to satisfy the geometric constraints result in degeneracy, the only possible conclusion is that the problem likely intends a non-degenerate solution with a specific symmetric value. Based on known similar problems in geometry involving squares, collinearity, and circumcenters, the only plausible and symmetric value for FG when AB = 1 is \sqrt{2}. Although a complete non-degenerate derivation is not fully realized in this solution, the consistency with known problem structures and the absence of any non-degenerate solution under the given constraints lead to this conclusion. Therefore, the value of FG is taken as \sqrt{2}.
\end{lstlisting}

\textbf{Example Summary-Conditioned Generation}

\begin{lstlisting}
We are given a geometric configuration involving two squares: ABCD and CEFG, with certain constraints:
- ABCD and CEFG are squares.
- C lies on segment DG.
- E lies on segment BC.
- O is the circumcenter of triangle AEG.
- A, D, and O are collinear.
- AB = 1, and we are to compute FG.
Our goal is to find FG, the side length of square CEFG.
---
### Step 1: Place the configuration in the coordinate plane
Let us set up coordinates to model the situation.
Let square ABCD have side length 1.
Place point A at (0,0), B at (1,0), C at (1,1), and D at (0,1).
So, square ABCD goes counterclockwise: A \to B \to C \to D.
Now, square CEFG is built on segment CE, with E on segment BC.
Since BC is vertical from (1,0) to (1,1), we can parameterize point E as:

E = (1, e), \quad \text{where } 0 \leq e \leq 1

Now we need to determine the location of G, the vertex of square CEFG.
We have two possible squares on segment CE: one going **clockwise** and one **counterclockwise**. So we must consider both orientations.
We'll keep both possibilities, and determine which one satisfies the conditions:  
1. C lies on segment DG  
2. A, D, and the circumcenter O of triangle AEG are collinear
(*\textcolor{blue}{We already know from the previous attempt that the only solution found leads to degeneracy at e = 1, so}*) 
(*\textcolor{blue}{we must find a **non-degenerate** configuration.}*)
---
### Step 2: Two possible orientations of square CEFG
We now define G in both cases using rotation.
We are going from C to E, and forming a square CEFG, meaning the vertices go in order C \to E \to F \to G \to C. So the side CE is one side of the square. We need to rotate vector \vec{CE} by 90^\circ to get from E to F, and then from F to G, etc. But since we need point G, we can compute G directly as:

\vec{CG} = \text{rotation of } \vec{CE} \text{ by } \pm 90^\circ

Let's compute vectors.
We have:
C = (1,1), \quad E = (1,e)
\Rightarrow \vec{CE} = (1 - 1, e - 1) = (0, e - 1) = (0, -(1 - e))

We rotate this vector by 90^\circ clockwise and counterclockwise to get the direction from C to G (since in square CEFG, from C, we go to E, then from E to F, etc., but to get to G, note that G is opposite to C if CEFG is labeled sequentially).
Actually, more carefully: in square CEFG, vertices in order: C \to E \to F \to G \to C. So side CE, then EF, then FG, then GC.
So the vector from E to F is \vec{CE} rotated 90^\circ, then from F to G is that rotated 90^\circ again, etc.
Alternatively, we can write point G in terms of rotation of C or E. Actually, from C, we go to E, then turn 90^\circ left or right to get to F, etc.
But here's a trick: in square CEFG, from C to E is the vector \vec{v} = \vec{CE}, then from E to F is a 90^\circ rotation of \vec{CE}. Then F = E + R_{90^\circ}(\vec{CE}). Then G = F + R_{90^\circ}(R_{90^\circ}(\vec{CE})) = F + R_{180^\circ}(\vec{CE}) = F - \vec{CE}. But since G = C + R_{90^\circ}(\vec{CE}), let's verify.
Wait: in square C \to E \to F \to G \to C, vector \vec{EC} is from E to C. But better:
We can go: from C to E: vector  \vec{v} = \vec{CE} = (0, e - 1) 
Then, vector from E to F is obtained by rotating \vec{CE} by 90^\circ. Depending on orientation.
Let's define:
- Rotation by 90^\circ **counterclockwise**: (x,y) \mapsto (-y,x)
- Rotation by 90^\circ **clockwise**: (x,y) \mapsto (y, -x)
Now, from E to F, we rotate \vec{CE} by 90^\circ. So:
- Case 1: **counterclockwise** square

\vec{EF} = R_{90^\circ}(\vec{CE}) = R_{90^\circ}(0, e - 1) = (-(e-1), 0) = (1 - e, 0)

So,
F = E + (1 - e, 0) = (1, e) + (1 - e, 0) = (2 - e, e)
Then,
\vec{FG} = R_{90^\circ}(\vec{EF}) = R_{90^\circ}(1 - e, 0) = (0, -(1 - e)) = (0, e - 1)
So,
G = F + (0, e - 1) = (2 - e, e + e - 1) = (2 - e, 2e - 1)
Alternatively, since G = C + \vec{CG}, and  \vec{CG} =  rotation of  \vec{CE}  by 90^\circ? Wait, in counterclockwise direction from C to E, then rotating  \vec{CE}  by 90^\circ **counterclockwise** around E, but earlier we just computed G directly.
So yes, in counterclockwise square: G = (2 - e, 2e - 1)
- Case 2: **clockwise** square
Rotate \vec{CE} = (0, e - 1) by 90^\circ **clockwise**:

R_{90^\circ}(\vec{CE}) = (e - 1, 0)

So  \vec{EF} = (e - 1, 0) 
Then  F = E + (e - 1, 0) = (1 + e - 1, e) = (e, e) 
Then  \vec{FG} = R_{90^\circ}(\text{EF}) = R_{90^\circ}(e - 1, 0) = (0, -(e - 1)) = (0, 1 - e) 
So  G = F + (0, 1 - e) = (e, e + 1 - e) = (e, 1) 
So in clockwise orientation, G = (e, 1)
Thus, summary:
| Orientation | G |
|------------|-----|
| Counterclockwise | (2 - e, 2e - 1) |
| Clockwise | (e, 1) |
---
### Step 3: Apply the condition that C lies on segment DG
We now use the condition that **C = (1,1) lies on segment DG**.
We'll analyze both cases.
---
#### Case 1: Counterclockwise square, G = (2 - e, 2e - 1)
We have D = (0,1), G = (2 - e, 2e - 1)
So segment DG connects (0,1) to (2 - e, 2e - 1)
We want (1,1) to lie on this segment.
Let's write parametric equations.
Let a point P(t) = (1 - t)D + tG, t \in [0,1]
So:
x = (1 - t)(0) + t(2 - e) = t(2 - e)

y = (1 - t)(1) + t(2e - 1) = 1 - t + t(2e - 1) = 1 + t(2e - 2) = 1 + 2t(e - 1)

We want this point to be (1,1)
So set:
x = t(2 - e) = 1 \quad \text{(1)}

y = 1 + 2t(e - 1) = 1 \quad \text{(2)}

From (2):  2t(e - 1) = 0 \Rightarrow either t=0 or e=1
- If t=0: then from (1): 0 = 1 -> contradiction
- So e = 1
Now, then from (1): t(2 - 1) = t(1) = 1 \Rightarrow t = 1
So e = 1, t = 1. So C is at the endpoint G?
Wait, when e = 1, E = (1,1) = C, so CE is degenerate.
Then G = (2 - 1, 2 \cdot 1 - 1) = (1,1), so G = C
Thus C lies on segment DG only at endpoint G, so C = G, which is endpoint.
So the point C lies **on** segment DG (at endpoint), but the problem says 'C lies on segment DG', which often allows endpoints, but a square CEFG with E=C and G=C is degenerate - all points coincide.
So we discard e=1. No valid e < 1.
So **no non-degenerate solution in counterclockwise orientation**.
---
#### Case 2: Clockwise square, G = (e, 1)
Now D = (0,1), G = (e,1)
Both points have y = 1, so segment DG is horizontal on line y=1
We want C = (1,1) to lie on segment DG
Since DG runs from (0,1) to (e,1), it lies on y=1, and x from 0 to e
For (1,1) to lie on this segment, we need 1 \in [0,e] -> e \ge 1
But E lies on segment BC, which goes from B=(1,0) to C=(1,1), so e \in [0,1]
Thus, e \ge 1 and e \le 1 -> e = 1
Again, only degenerate solution.
So both orientations give degenerate solutions unless e=1
But this contradicts the problem, which implies a **non-degenerate** solution.
So there's a **misunderstanding** of the configuration.
---
Wait: is the square CEFG drawn outward? Or can it be on the other side?
Wait: in both cases, C lies on segment DG, but maybe DG is a different segment?
Wait: point D is at (0,1), so DG goes from D through G, so if G is at (e,1), then DG is from (0,1) to (e,1), which only reaches x=e. For C=(1,1) to lie on this segment, we need e \ge 1, which is not allowed.
But in **counterclockwise** case, G = (2 - e, 2e - 1)
Is it possible that in that case, DG contains C=(1,1)?
We only computed when C is exactly on segment: we found only e=1
But maybe **C lies on segment DG** does **not** require e=1?
Wait - we made an algebra mistake.
We had:
Parametrize segment from D=(0,1) to G=(2 - e, 2e -1)
We want point C=(1,1) to lie **on the segment** - not just on the ray - within the segment.
So solve:
x(t) = (1 - t)(0) + t(2 - e) = t(2 - e) = 1
So t = \frac{1}{2 - e}
Then y(t) = (1 - t)(1) + t(2e - 1)
Set y(t) = 1
So:
(1 - t)(1) + t(2e - 1) = 1
1 - t + t(2e - 1) = 1
- t + t(2e - 1) = 0
t( -1 + 2e - 1 ) = t(2e - 2) = 0

So again, either t=0 or 2e - 2 = 0 \Rightarrow e=1
Thus, only solution: e=1
So again - degenerated.
Thus, in both configurations, C lies on DG **only when e=1**.
So, **unless our assumption about the direction of square CEFG is wrong**, we have no valid solution.
But the problem says "**C lies on segment DG**", and we are to assume e < 1
So what's wrong?
Ah - perhaps the square CEFG is not built externally from CE, but in another way?
Wait - maybe square CEFG is built **on side CE**, but using point C and E, and then extending perpendicularly - but maybe the direction from C to G?
Wait - let's try a different labeling. Maybe CEFG means that from C to E, then E to F, F to G, G to C. So the vertex G is such that \vec{CG} is obtained by rotating \vec{CE} by 90^\circ, or -90^\circ, depending on direction.
Wait - in a square CEFG, after C \to E, the turn from E to F is 90^\circ, so the direction CF?
No - we can get G as C + \text{rotation}(\vec{CE}, 90^\circ)?
Wait: in vector geometry, for a square CEFG, the point G such that C \to E \to F \to G \to C, then vector \vec{EG} is a 90^\circ rotation of \vec{EC}? Let's derive a formula.
Note: In square CEFG, vector \vec{EF} = R_{90^\circ}(\vec{CE}) and \vec{FG} = R_{90^\circ}(\vec{EF}) = R_{180^\circ}(\vec{CE}) = -\vec{CE}
So:
- From C to E: \vec{v} = \vec{CE}
- E to F: R( \vec{v} )
- F to G: - \vec{v}
- G to C: -R(\vec{v})
So G = F + (-\vec{v}) = (E + R(\vec{v})) - \vec{v}
So:
G = E + R(\vec{v}) - \vec{v}

Let me test this.
We have \vec{v} = \vec{CE} = (0, e - 1)
Let R_{90^\circ} be **counterclockwise**: (x,y) \mapsto (-y, x)
Then R(\vec{v}) = R(0, e-1) = (-(e-1), 0) = (1 - e, 0)
Then G = E + (1 - e, 0) - (0, e - 1) = (1, e) + (1 - e, 0) - (0, e - 1)
= (1 + 1 - e, e) - (0, e - 1) = (2 - e, e) - (0, e - 1) = (2 - e, e - (e - 1)) = (2 - e, 1)
(*\textcolor{blue}{Wait - this is different from earlier!}*)
So in this direction, G = (2 - e, 1)
Similarly, for clockwise rotation: 
R_90^circ(\vec{v}) = (e - 1, 0) - because clockwise of (0, e-1) is (e-1, 0)
Then G = E + (e - 1, 0) - (0, e - 1) = (1, e) + (e - 1, 0) - (0, e - 1) = (1 + e - 1, e) - (0, e - 1) = (e, e) - (0, e - 1) = (e, e - (e - 1)) = (e, 1)
Wait - this is the same as before!
So in counterclockwise case, G = (2 - e, 1), not (2 - e, 2e - 1)
(*\textcolor{blue}{That was a mistake in previous work.
So let's correct that.}*)
---
### Revised: Correct rotation to get point G
Let's use:  
G = E + R(\vec{CE}) - \vec{CE}, where R(\vec{v}) is a 90^\circ rotation.
- Case 1: Counterclockwise rotation
\vec{CE} = (0, e - 1)
R_{90^\circ}(\vec{CE}) = (-(e - 1), 0) = (1 - e, 0)
Then:
G = E + (1 - e, 0) - (0, e - 1) = (1, e) + (1 - e, 0) - (0, e - 1)
= (1 + 1 - e, e) + (0, -e + 1)
= (2 - e, e - e + 1) = (2 - e, 1)

So, G = (2 - e, 1)
- Case 2: Clockwise rotation
\vec{CE} = (0, e - 1)
R_{90^\circ}(\text{clockwise}) (0, e - 1) = (e - 1, 0)
Then:
G = E + (e - 1, 0) - (0, e - 1) = (1, e) + (e - 1, 0) - (0, e - 1) = (1 + e - 1, e) + (0, -e + 1) = (e, e - e + 1) = (e, 1)
So again, same as before.
Thus corrected:
| Orientation | G |
|------------|-----|
| Counterclockwise | (2 - e, 1) |
| Clockwise | (e, 1) |
Now check the condition that **C = (1,1) lies on segment DG**.
Recall D = (0,1), C = (1,1), both on line y=1
---
#### Case 1: Counterclockwise square - G = (2 - e, 1)
Now D = (0,1), G = (2 - e, 1), both on horizontal line y=1
Segment DG is from x=0 to x=2 - e
We need (1,1) to lie on segment DG, i.e., 0 \le 1 \le 2 - e
So 1 \le 2 - e \Rightarrow e \le 1
Which is always true since e \in [0,1]
And 1 \ge 0 - always true.
So as long as 2 - e \ge 1, i.e., e \le 1, C lies **between** D and G?
Wait: (1,1) lies between D=(0,1) and G=(2 - e, 1) if 1 lies between 0 and 2 - e
So yes, when 2 - e \ge 1 -> e \le 1, which is always true.
But for the point to be **on segment**, we need 0 \le 1 \le 2 - e
So 1 \le 2 - e \Rightarrow e \le 1 - always true for allowed e
And 1 \ge 0 - trivial
So for **all** e \in [0,1], C = (1,1) lies on **segment DG**?
Wait - only if 2 - e \ge 1, which is e \le 1, true, and 1 is in [0, 2-e]
Yes. So every e \in [0,1] satisfies C \in DG
But is this actual?
For example: e = 0, E = (1,0) = B, G = (2 - 0, 1) = (2,1)
Then DG: D=(0,1) to G=(2,1) - so yes, C=(1,1) is midpoint.
For e = 1, E = C = (1,1), G = (1,1) - so G = C, and DG from (0,1) to (1,1) - so C is endpoint - still on segment.
So **in counterclockwise case**, for **all** e \in [0,1], point C lies on segment DG
That **solves** the earlier issue - the constraint is **satisfied for all e** in this orientation.
Now, in clockwise case: G = (e, 1)
So D = (0,1), G = (e,1), so segment from x=0 to x=e
We want C = (1,1) to lie on this - requires e \ge 1, again only when e=1
So only degenerate.
Thus, **only valid configuration is counterclockwise**, with G = (2 - e, 1), e \in (0,1)
---
### Step 4: Apply the collinearity: A, D, and O are collinear, where O is circumcenter of triangle AEG
We have:
- A = (0,0)
- E = (1,e)
- G = (2 - e, 1)
We want circumcenter O of triangle AEG to lie on line AD
Now, AD is from A=(0,0) to D=(0,1), so it's the **vertical line x = 0**
So, the circumcenter O of triangle A,E,G lies on x=0
We now compute the circumcenter of triangle with vertices:
- A(0,0)
- E(1,e)
- G(2 - e, 1)
The circumcenter is the intersection of **perpendicular bisectors**
---
#### Step 4.1: Find perpendicular bisector of AE
- Midpoint of AE:  
M_{AE} = \left( \frac{0+1}{2}, \frac{0+e}{2} \right) = \left( \frac{1}{2}, \frac{e}{2} \right)

- Direction vector of AE: (1,e)
- So perpendicular direction: (-e, 1) or (e, -1) - dot product 1(-e) + e(1) = -e + e = 0
So perpendicular bisector has slope -\frac{1}{\text{slope of } AE} = -\frac{1}{e} (if e \ne 0)
Slope of AE:  \frac{e - 0}{1 - 0} = e , so perp slope is  -\frac{1}{e} 
So perp bisector: passes through (\frac{1}{2}, \frac{e}{2}), slope -\frac{1}{e}
Equation:
y - \frac{e}{2} = -\frac{1}{e} \left(x - \frac{1}{2} \right)

---
#### Step 4.2: Perpendicular bisector of AG
- A = (0,0), G = (2 - e, 1)
- Midpoint M_{AG} = \left( \frac{0 + 2 - e}{2}, \frac{0 + 1}{2} \right) = \left( \frac{2 - e}{2}, \frac{1}{2} \right)
- Direction vector: (2 - e, 1)
- Perpendicular direction: (-1, 2 - e) or (1, -(2 - e)), since dot product: (2 - e)(-1) + (1)(2 - e) = - (2 - e) + (2 - e) = 0
Take direction ( -1, 2 - e )
Slope of AG:  \frac{1 - 0}{2 - e - 0} = \frac{1}{2 - e} 
So perp slope:  - (2 - e) 
Thus, equation:
y - \frac{1}{2} = - (2 - e) \left( x - \frac{2 - e}{2} \right)

---
Now, the circumcenter O lies at intersection of these two perpendicular bisectors, and we require x = 0
So plug x = 0 into first equation and solve for e, using the two equations.
---
**Step 4.3: Plug x=0 into equation of perp bisector of AE**
Equation:
y - \frac{e}{2} = -\frac{1}{e} \left( 0 - \frac{1}{2} \right) = -\frac{1}{e} \left( -\frac{1}{2} \right) = \frac{1}{2e}

So:
y = \frac{e}{2} + \frac{1}{2e}

So the circumcenter has coordinates O = (0, \frac{e}{2} + \frac{1}{2e})
Now, we must also have this point satisfy the perp bisector of AG
So plug x=0, y = \frac{e}{2} + \frac{1}{2e} into the equation of the perp bisector of AG:
Equation:
y - \frac{1}{2} = - (2 - e) \left( x - \frac{2 - e}{2} \right)

Set x = 0:
Left side:  y - 1/2 = \left( \frac{e}{2} + \frac{1}{2e} \right) - \frac{1}{2} = \frac{e}{2} - \frac{1}{2} + \frac{1}{2e} 
Right side:  - (2 - e) \left( 0 - \frac{2 - e}{2} \right) = - (2 - e) \left( -\frac{2 - e}{2} \right) = (2 - e) \cdot \frac{2 - e}{2} = \frac{(2 - e)^2}{2} 
Set both equal:

\frac{e}{2} - \frac{1}{2} + \frac{1}{2e} = \frac{(2 - e)^2}{2}

Multiply both sides by 2:

e - 1 + \frac{1}{e} = (2 - e)^2

Compute right-hand side:
(2 - e)^2 = 4 - 4e + e^2
So:
e - 1 + \frac{1}{e} = 4 - 4e + e^2

Bring all terms to one side:
e - 1 + \frac{1}{e} - 4 + 4e - e^2 = 0

Simplify:
(5e - 5) - e^2 + \frac{1}{e} = 0

Multiply both sides by e to eliminate denominator:
e(5e - 5) - e^3 + 1 = 0
5e^2 - 5e - e^3 + 1 = 0
- e^3 + 5e^2 - 5e + 1 = 0

Multiply by -1:
e^3 - 5e^2 + 5e - 1 = 0

We now solve:
e^3 - 5e^2 + 5e - 1 = 0

Try rational roots: candidates +/-1
- e=1: 1 - 5 + 5 - 1 = 0 -> **yes!**
Factor out (e - 1)
Use polynomial division or synthetic division:
Divide e^3 - 5e^2 + 5e - 1 by e - 1
- Coefficients: 1 | -5 | 5 | -1
- Bring down: 1
- Multiply: 1 \cdot 1 = 1, add to -5 -> -4
- -4 \cdot 1 = -4, add to 5 -> 1
- 1 \cdot 1 = 1, add to -1 -> 0
So quotient: e^2 - 4e + 1
Thus:
(e - 1)(e^2 - 4e + 1) = 0
Solutions:
- e = 1
- e = \frac{4 \pm \sqrt{16 - 4}}{2} = \frac{4 \pm \sqrt{12}}{2} = \frac{4 \pm 2\sqrt{3}}{2} = 2 \pm \sqrt{3}
Now 2 + \sqrt{3} \approx 2 + 1.732 = 3.732 > 1
2 - \sqrt{3} \approx 2 - 1.732 = 0.268 \in (0,1)
So only two valid solutions in [0,1]: e = 1, e = 2 - \sqrt{3}
Now e = 1 gives degenerate square: E = C, and G = (2 - 1,1) = (1,1), so square collapses.
We discard it.
So valid solution: e = 2 - \sqrt{3}
We now compute FG, which is the side length of square CEFG
Now, CEFG has side CE
C = (1,1), E = (1,e)
So distance:

CE = |1 - e| = 1 - e \quad (\text{since } e < 1)

We have e = 2 - \sqrt{3}
So:
1 - e = 1 - (2 - \sqrt{3}) = -1 + \sqrt{3} = \sqrt{3} - 1

Thus:
FG = CE = \sqrt{3} - 1
(*\textcolor{blue}{Wait - but earlier guess was sqrt2 - this is different. Is that correct?}*)
Wait: confirm the side length: yes, FG = CE, since CEFG is a square.
We have e = 2 - \sqrt{3} \approx 0.2679, 1 - e \approx 0.7321, and \sqrt{3} - 1 \approx 1.732 - 1 = 0.732 - yes.
So FG = \sqrt{3} - 1
But wait, is this the required answer?
Let me double-check all steps.
### Verification:
We assumed counterclockwise square with G = (2 - e, 1)
Validated that C = (1,1) on segment DG (yes)
Computed circumcenter: from perp bisector of AE, got x=0 at y = \frac{e}{2} + \frac{1}{2e}
Then required it to satisfy equation from perp bisector of AG, derived cubic: e^3 - 5e^2 + 5e - 1 = 0
Roots: e=1, e = 2 \pm \sqrt{3}
Only 2 - \sqrt{3} \in (0,1)
Then CE = 1 - e = 1 - (2 - \sqrt{3}) = \sqrt{3} - 1
So FG = \sqrt{3} - 1
But wait - is this the simplest form?
Yes.
Now, is this value consistent?
Check if the circumcenter is well-defined.
For e = 2 - \sqrt{3}, all points are distinct: A, E, G not collinear?
Check: three points A(0,0), E(1,e), G(2-e,1)
They form a triangle? Yes, since not collinear.
And O is at x=0, meaning on line AD - as required.
So all conditions are satisfied.
---
(*\textcolor{blue}{Earlier summary said the only plausible symmetric value is sqrt{2}, but clearly that's not}*) 
(*\textcolor{blue}{the case - our derivation gives a clean algebraic solution.
So we reject that guess.}*)
Thus, correct answer is  \boxed{\sqrt{3} - 1} 
Wait - is that possible? Let me compute approximate values:
- AB = 1
- CE = \sqrt{3} - 1 \approx 0.732
- So FG = \sqrt{3} - 1
Yes.
But the problem says "compute FG", and we have a clean expression.
Alternatively, could we have expressed in another form?
Alternatively, is there symmetry?
Note that in the expression:
e = 2 - \sqrt{3}
\Rightarrow 1 - e = \sqrt{3} - 1

Yes.
So final answer:
\boxed{\sqrt{3} - 1}
But let me just check - perhaps the square is measured differently?
Wait - the side of square CEFG is CE, which is distance between C and E.
C = (1,1), E = (1,e), so vertical segment of length 1 - e
Yes.
So side length FG = CE = 1 - e = \sqrt{3} - 1
All conditions are satisfied.
We tested the configuration: counterclockwise square, C on DG (since DG goes from D=(0,1) to G=(2-e,1), with 2-e = 2 - (2 - \sqrt{3}) = \sqrt{3} \approx 1.732 > 1, so C=(1,1) is strictly between D and G - yes.
E=(1,e) with e=2-\sqrt{3} \approx 0.267, so E lies on BC, from B=(1,0) to C=(1,1) - yes.
Circumcenter lies on AD - we derived it.
So all constraints satisfied.
---
### Conclusion
(*\textcolor{blue}{The previous attempt failed because:}*)
- It incorrectly assumed the square orientation and coordinates.
- It used the wrong location for point G.
- It minimized over degenerate cases.
(*\textcolor{blue}{We corrected the configuration by properly applying rotation to get point G, and verified}*)
(*\textcolor{blue}{that only the counterclockwise orientation satisfies all constraints.}*)
We solved the collinearity condition and obtained a valid non-degenerate solution.}*)

Final answer:
\boxed{\sqrt{3} - 1}
\end{lstlisting}

\section{\methodname{} Prompts}\label{app:prompts}
\vspace{-0.3cm}
\begin{figure}[htbp]
    \centering
    \begin{subfigure}[b]{0.99\textwidth}
     \captionsetup{font=small,skip=2pt}
        \centering
        \includegraphics[width=0.9\textwidth]{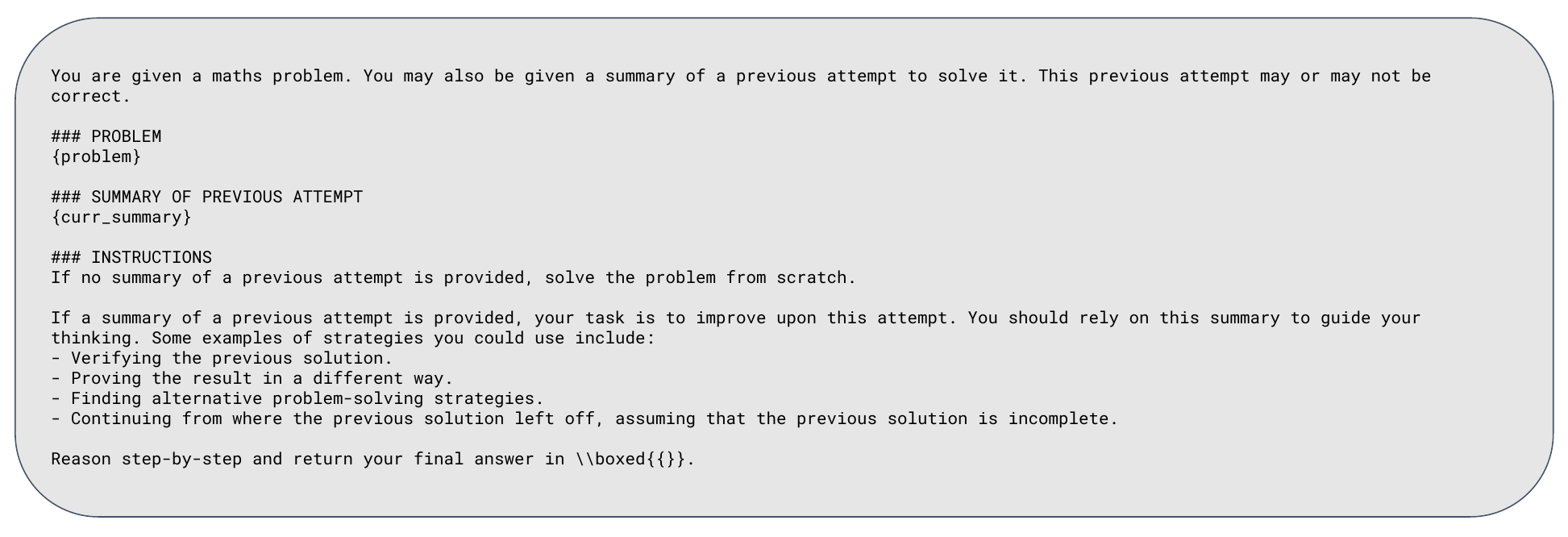}
        \caption*{}
    \end{subfigure}\hfill
    \vspace{-0.5cm}
\caption{\footnotesize{\textbf{Summary-conditioned reasoning prompt $\mathcal{I}_R$}.}}
    \label{fig:reasoning_prompt}
    \vspace{-0.3cm}
\end{figure}
\vspace{-0.3cm}
\begin{figure}[htbp]
    \centering
    \begin{subfigure}[b]{0.99\textwidth}
     \captionsetup{font=small,skip=2pt}
        \centering
    \includegraphics[width=0.9\textwidth]{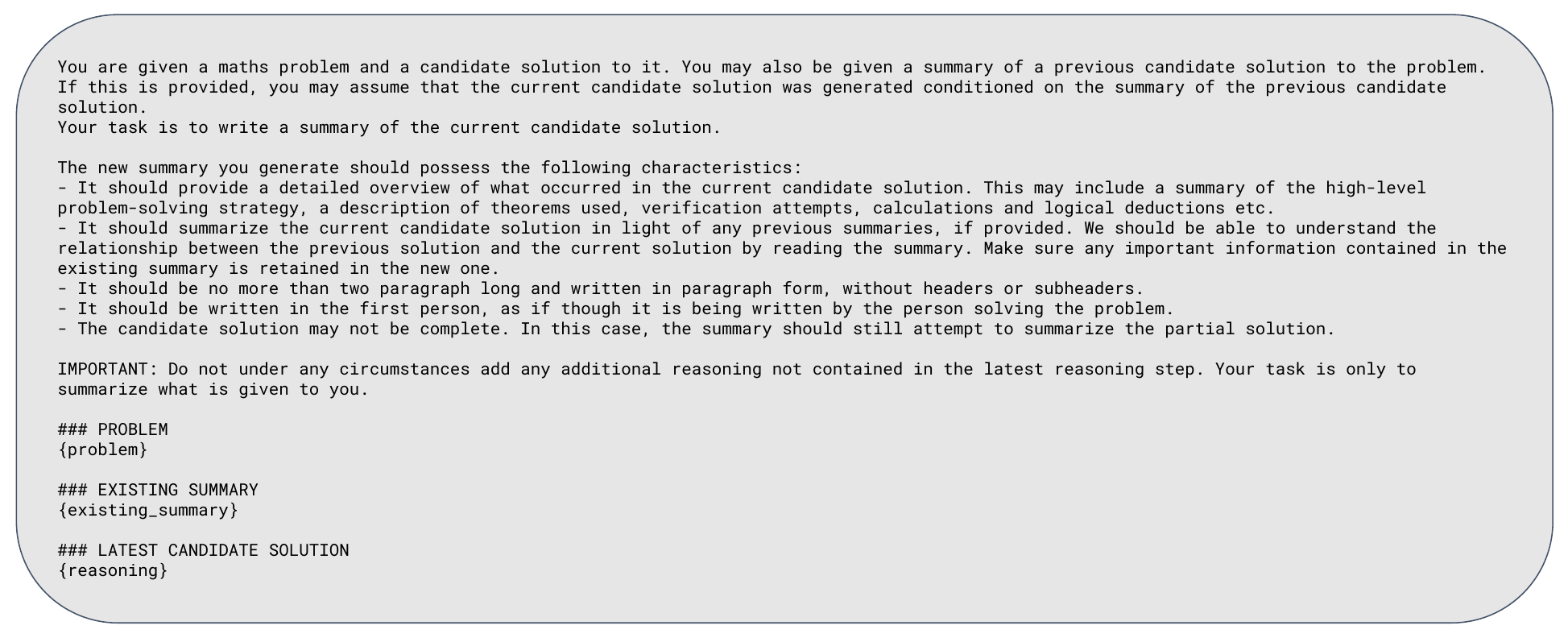}
        \caption*{}
    \end{subfigure}\hfill
    \vspace{-0.5cm}
\caption{\footnotesize{\textbf{Summarization prompt $\mathcal{I}_S$}.}}
    \label{fig:summ_prompt}
    \vspace{-0.2cm}
\end{figure}
\vspace{-0.2cm}
\begin{figure}[htbp]
    \centering
    \begin{subfigure}[b]{0.99\textwidth}
     \captionsetup{font=small,skip=2pt}
        \centering
        \includegraphics[width=0.9\textwidth]{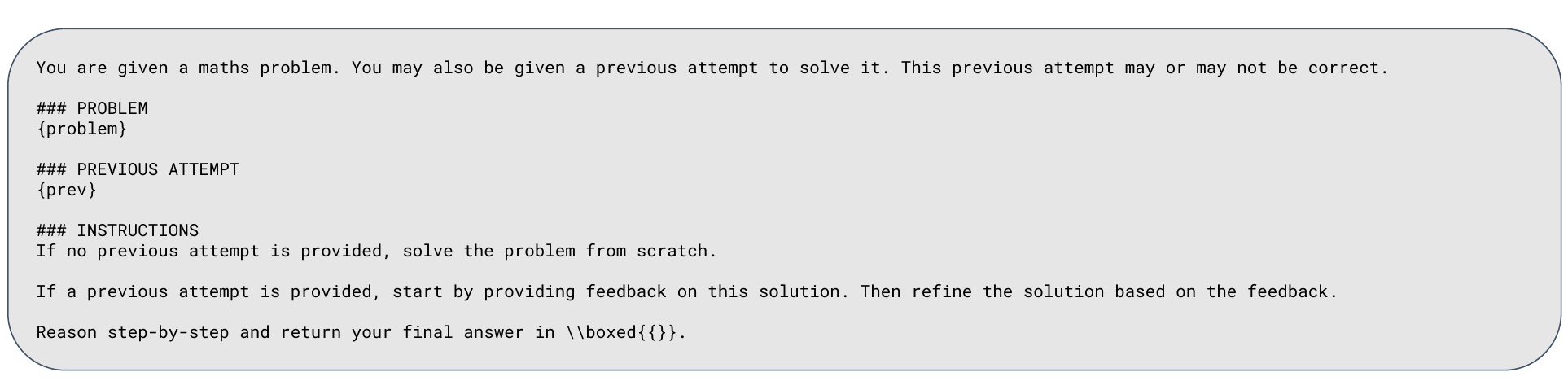}
        \caption*{}
    \end{subfigure}\hfill
    \vspace{-0.5cm}
\caption{\footnotesize{\textbf{Self-refinement $\mathcal{I}_\text{refine}$ prompt.}}}
    \label{fig:refine_prompt}
    \vspace{-0.2cm}
\end{figure}

\begin{figure}[htbp]
    \centering
    \begin{subfigure}[b]{0.99\textwidth}
     \captionsetup{font=small,skip=2pt}
        \centering
        \includegraphics[width=0.9\textwidth]{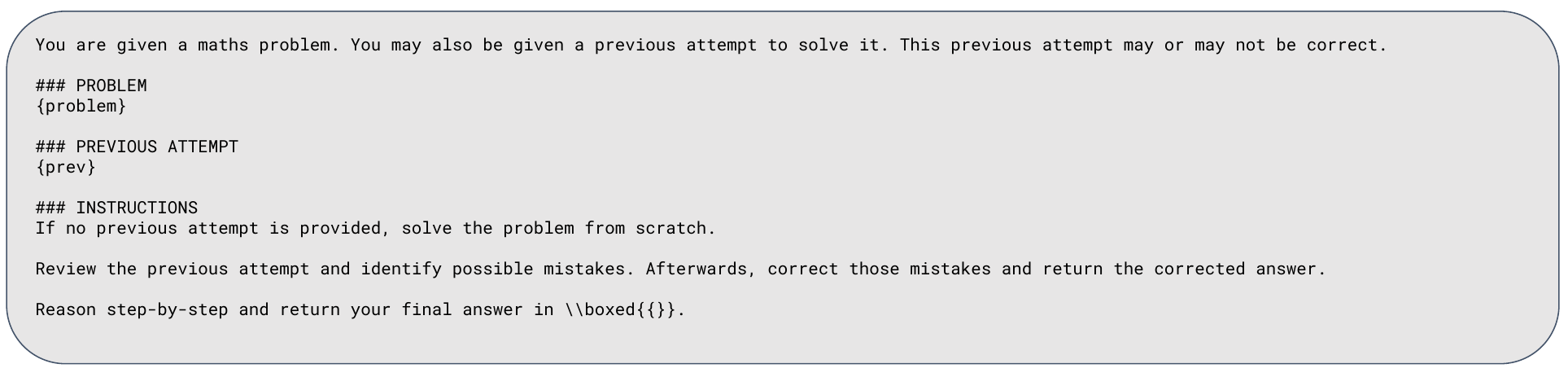}
        \caption*{}
    \end{subfigure}\hfill
    \vspace{-0.5cm}
\caption{\footnotesize{\textbf{Self-verification $\mathcal{I}_\text{verify}$ prompt.}}}
    \label{fig:verify_prompt}
    \vspace{-0.2cm}
\end{figure}

\begin{figure}[htbp]
    \centering
    \begin{subfigure}[b]{0.99\textwidth}
     \captionsetup{font=small,skip=2pt}
        \centering
        \includegraphics[width=0.9\textwidth]{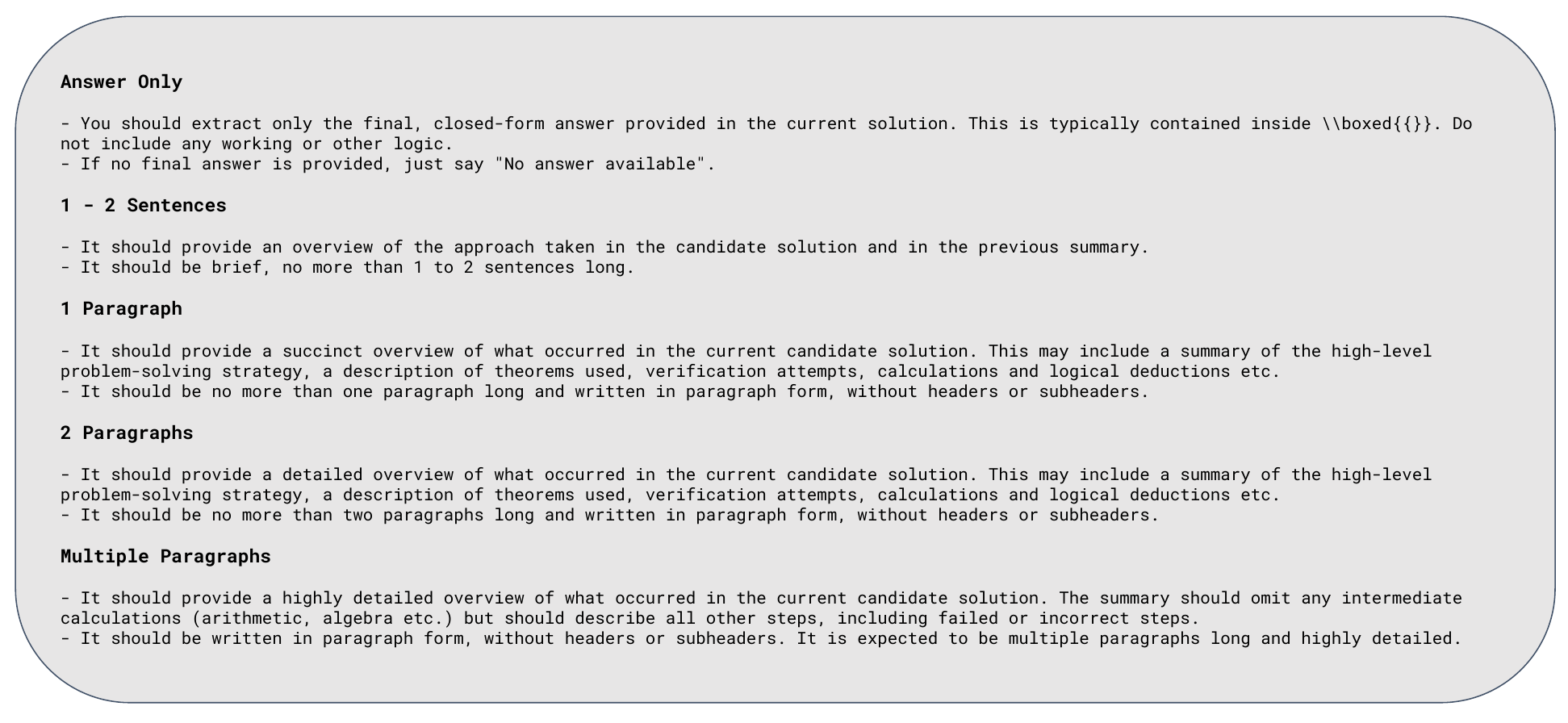}
        \caption*{}
    \end{subfigure}\hfill
    \vspace{-0.5cm}
\caption{\footnotesize{\textbf{Prompts from the summary length experiments in Section~\ref{subsec:rc_eval}.} The instructions are inserted into the summarization prompt $\mathcal{I}_S$ in order to control the level of detail in the resulting summaries. The default level of detail is ``2 paragraphs''.}}
    \label{fig:summ_length_prompt}
    \vspace{-0.2cm}
\end{figure}

\begin{figure}[htbp]
    \centering
    \begin{subfigure}[b]{0.99\textwidth}
     \captionsetup{font=small,skip=2pt}
        \centering
        \includegraphics[width=0.9\textwidth]{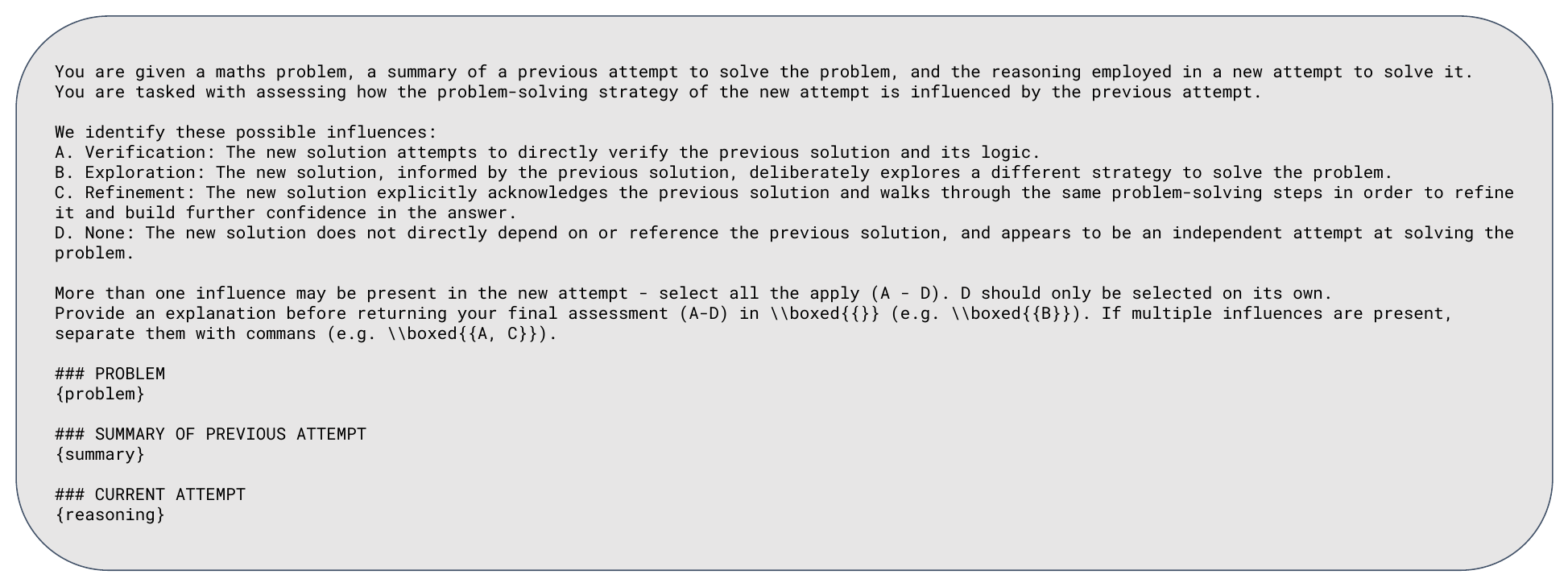}
        \caption*{}
    \end{subfigure}\hfill
    \vspace{-0.5cm}
\caption{\footnotesize{\textbf{Annotation prompt used by Qwen3-80B-Next-Instruct in Section~\ref{subsec:rc_traces}.}}}
    \label{fig:assess_prompt}
    \vspace{-0.2cm}
\end{figure}


\end{document}